%% file: Main.tex
\documentclass[conference,compsoc]{IEEEtran}
\usepackage{amsmath}
\usepackage{algorithmic}
\usepackage{graphicx}
\usepackage{textcomp}
\usepackage{xcolor}
\usepackage{pifont}
\usepackage{multirow}
\usepackage{amssymb}
\usepackage{makecell}
\usepackage{color}
\usepackage{colortbl}
\usepackage{booktabs}
\usepackage{enumitem}
\usepackage{soul}

\usepackage{url}
 
\usepackage[
  breaklinks = true,
  unicode = true,
  urlcolor = blue,
  colorlinks = true,
  citecolor = blue,
  linkcolor = blue ]{hyperref}

\definecolor{tabcomparegarycolor}{RGB}{245,245,245}

\usepackage{caption} 

\newif\ifshepherd

\shepherdfalse

\ifshepherd
\newcommand{\hjj}[1]{\textcolor{blue}{#1}}

\newcommand{\lb}[1]{\textcolor{blue}{#1}}
\newcommand{\clrrow}[1]{\rowcolor{#1}}
\newcommand{\clrcolumn}[1]{\columncolor{#1}}
\newcommand{\clrcell}[1]{\cellcolor{#1}}

\newcommand{\revbegin}[2]{\textcolor{red}{[R#1-Q#2 Begin:]}}
\newcommand{\revend}[2]{\textcolor{red}{[:End R#1-Q#2]}}
\newcommand{\conbegin}[1]{\textcolor{red}{[Concern-#1 Begin:]}}
\newcommand{\conend}[1]{\textcolor{red}{[:End Concern-#1]}}
\else 
\newcommand{\hjj}[1]{#1}

\newcommand{\lb}[1]{#1}
\newcommand{\clrrow}[1]{}
\newcommand{\clrcolumn}[1]{}
\newcommand{\clrcell}[1]{}

\newcommand{\revbegin}[2]{}
\newcommand{\revend}[2]{}
\newcommand{\conbegin}[1]{}
\newcommand{\conend}[1]{}
\fi

\ifCLASSOPTIONcompsoc
  \usepackage[nocompress]{cite}
\else
  \usepackage{cite}
\fi

\setstcolor{blue}
\soulregister{\cite}{7}

\begin{document}
\IEEEoverridecommandlockouts
%
% paper title
% Titles are generally capitalized except for words such as a, an, and, as,
% at, but, by, for, in, nor, of, on, or, the, to and up, which are usually
% not capitalized unless they are the first or last word of the title.
% Linebreaks \\ can be used within to get better formatting as desired.
% Do not put math or special symbols in the title.
% \title{Bare Demo of IEEEtran.cls for\\ IEEE Computer Society Conferences}

\title{Fight Fire with Fire: Combating Adversarial Patch Attacks using Pattern-randomized Defensive Patches}
% \author{\IEEEauthorblockN{Anonymous Authors}}

\author{
\IEEEauthorblockN{Jianan Feng, Jiachun Li, Changqing Miao, Jianjun Huang, Wei You, Wenchang Shi and Bin Liang\thanks{\textnormal{*Bin Liang is the corresponding author.}}\IEEEauthorrefmark{1}}
\IEEEauthorblockA{School of Information, Renmin University of China, Beijing, China}
\IEEEauthorblockA{\{jiananfeng, jclee, miaochangqing, hjj, youwei, wenchang, liangb\}@ruc.edu.cn}
}

\maketitle
% 1001-添加页眉页脚
% \thispagestyle{fancy} % IEEE模板在\maketitle后会自动声明\thispagestyle{plain}，
%                     % 导致第一页什么都没有。所以得把plain更改为fancy
% \lhead{} % 页眉左，需要东西的话就在{}内添加
% \chead{} % 页眉中
% \rhead{} % 页眉右
% \lfoot{} % 页脚左
% \cfoot{\thepage} % 页脚中，\thepage 表示当前页码
% \rfoot{} %页脚右
% \renewcommand{\headrulewidth}{0pt} %改为0pt即可去掉页眉下面的横线
% \renewcommand{\footrulewidth}{0pt} %改为0pt即可去掉页脚上面的横线

% As a general rule, do not put math, special symbols or citations
% in the abstract
\begin{abstract}
% The abstract goes here.
Object detection has found extensive applications in various tasks, but it is also susceptible to adversarial patch attacks.
% Existing defense methods often necessitate modifications to the target model or result in unacceptable time overhead. In this paper, we adopt a counterattack approach, following the principle of ``fight fire with fire'', and propose a novel and general methodology for defending adversarial attacks. We utilize an active defense strategy by injecting two types of defensive patches, \textit{canary} and \textit{woodpecker}, into the input to proactively probe or counteract potential adversarial patches without altering the target model.
The ideal defense should be effective, efficient, easy to deploy, and capable of withstanding adaptive attacks.
In this paper, we adopt a counterattack strategy to propose a novel and general methodology for defending adversarial attacks. Two types of defensive patches, \textit{canary} and \textit{woodpecker}, are specially-crafted and injected into the model input to proactively probe or counteract potential adversarial patches.
In this manner, adversarial patch attacks can be effectively detected by simply analyzing the model output, without the need to alter the target model.
Moreover, we employ \textit{randomized} canary and woodpecker injection patterns to defend against defense-aware attacks. The effectiveness and practicality of the proposed method are demonstrated through comprehensive experiments. The results illustrate that canary and woodpecker achieve high performance, even when confronted with unknown attack methods, while incurring limited time overhead. Furthermore, our method also exhibits sufficient robustness against defense-aware attacks, as evidenced by adaptive attack experiments.
\end{abstract}

% no keywords

% For peer review papers, you can put extra information on the cover
% page as needed:
% \ifCLASSOPTIONpeerreview
% \begin{center} \bfseries EDICS Category: 3-BBND \end{center}
% \fi
%
% For peerreview papers, this IEEEtran command inserts a page break and
% creates the second title. It will be ignored for other modes.
\IEEEpeerreviewmaketitle

\input{Introduction}

\input{Background}

\input{Approach}

\input{Evaluation}

\input{Discussion}

\input{Related}

\input{Conclusion}

\section*{Availability}
The source code and dataset are available at  \url{https://github.com/1j4j1230/Fight_Fire_with_Fire}. 
% \textcolor{red}{\bf \hl{The source code will be released upon publication.}}

\section*{ACKNOWLEDGMENTS}
The authors would like to thank the reviewers for their constructive comments. 
The work is supported in part by National Natural Science Foundation of China (NSFC) under grants 62272465, 62441206, 62002361, and 62272464,
and the CCF-Huawei Populus Grove Fund under grant CCF-HuaweiSE202310, 
and the Fundamental Research Funds for
the Central Universities and the Research Funds of Renmin University of China under grant 22XNKJ29.

\bibliographystyle{IEEEtranS}
\bibliography{IEEEabrv,Citation}

\input{Appendix}
\input{Meta-review}

\end{document}

%% file: Introduction.tex
\section{Introduction}

In recent years, modern object detection models have demonstrated exceptional performance and are widely utilized in various physical-world tasks, such as pedestrian detection \cite{Huang20pedestrian,Kim22pedestrian,pang2019mask} and autonomous driving \cite{Chen17autodriving,Zheng2021autodriving}. Unfortunately, they have been proven to be vulnerable to adversarial patch attacks, involving hiding attacks \cite{attack_fool,attack_advtext} and misclassification attacks \cite{attack_naturalistic,attack_upc}. Clearly, how to safeguard object detectors against adversarial patch attacks is an important and urgent problem. 

Several defense methods ~\cite{DefendEnergy, DefenseMask, LiuSegDefend, ObjectSeeker, DongRobustDet, DetectorGuard} have been proposed. Unfortunately, the existing methods often require modification of the target model, such as introducing new net layers ~\cite{DefenseMask,LiuSegDefend}, adding a new model ~\cite{DetectorGuard}, or retraining ~\cite{DongRobustDet}.
% It is impractical for most physical-world scenarios.
In addition, the time overhead of SOTA methods, such as ~\cite{ObjectSeeker}, may increase significantly. 
% For example, ObjectSeeker ~\cite{ObjectSeeker} can lead to an increase in detection time of 3,900\% on YOLOv2.
Furthermore, it is crucial to acknowledge that attackers consistently maintain the upper hand relative to defenders. In many cases, attackers can gain knowledge about the target model and defense techniques through means such as reverse engineering, enabling them to launch defense-aware attacks, as demonstrated in ~\cite{DBLP:conf/nips/TramerCBM20, www16, GoodFishman}. 
% In practice, existing attack methods involve iteratively optimizing adversarial patches on the target model.
Attackers can adjust their optimization objectives to generate adaptive adversarial patches that can bypass the defense mechanisms.
% Therefore, an effective defense method should possess high detection capabilities while also being easily deployable, low in computational overhead, and sufficiently robust against defense-aware attacks.
Therefore, an effective defense method should possess high detection capabilities, ideally without altering the target model, low computational costs, and sufficient robustness against defense-aware attacks.

% In this study, we employ the same way of adversarial attacks to counter them (\textit{fighting fire with fire}). Adversarial patches, fundamentally, are additional elements introduced by attackers into the model's input to perturb nearby objects, resulting in their misidentification or undetection. In like manner, some specially designed elements can also be introduced into the input to sense or perturb adversarial patches.

\begin{figure}[t]
\centering
\includegraphics[width=\linewidth]{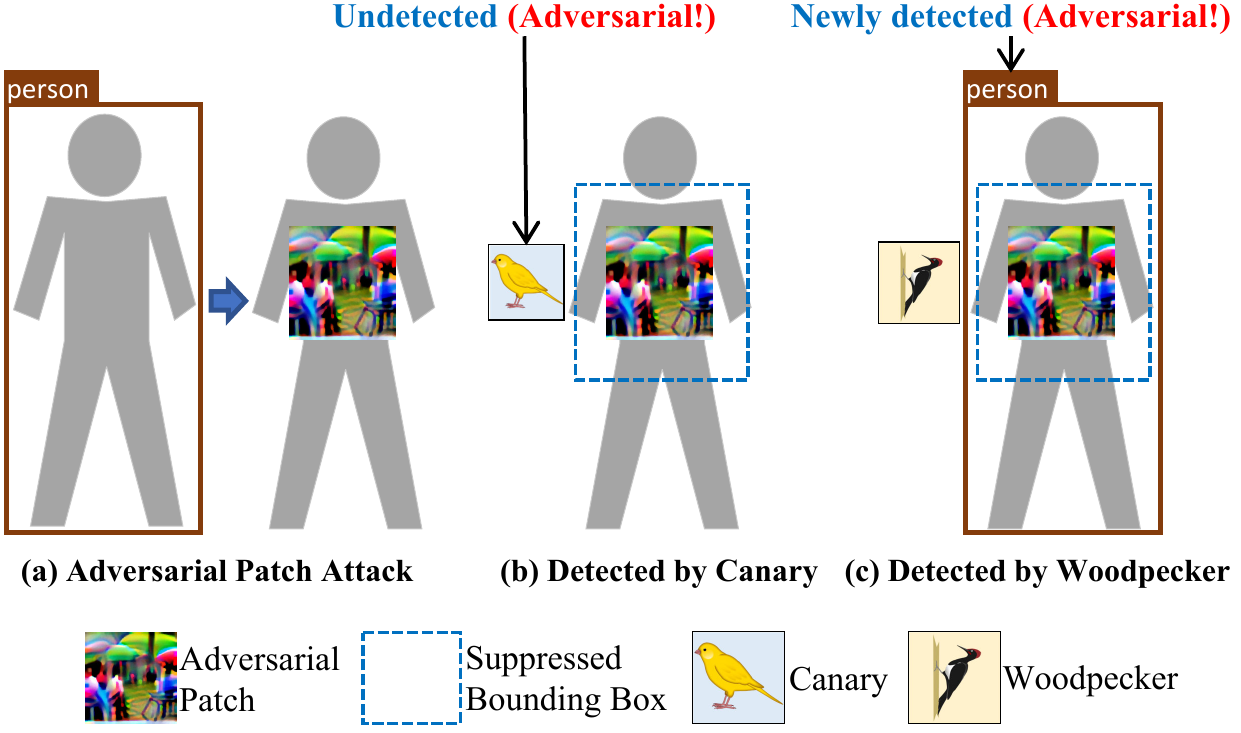}
\caption{Basic idea of canary and woodpecker.}
\label{fig-idea}
\end{figure}

In this study, we employ the same way of adversarial attacks to counter them. Adversarial patches are essentially additional elements introduced by attackers into the model input to perturb target objects, resulting in their misidentification or disappearance (Figure~\ref{fig-idea}(a)). In a similar manner, we can also proactively introduce specially designed elements, termed \textbf{defensive patches}, into the input to defend against adversarial patch attacks (\textit{fight fire with fire}).

% % Based on the above idea, this paper proposes a novel and general methodology to defend adversarial patch attacks. Rather than passively analyzing the original input, we adopt an active defense strategy by proactively injecting new elements, termed \textit{defensive patches}, into the input. In this way, potential attacks can be detected by examining the changes in the model output after incorporating defensive patches. Our patches are generated to probe or weaken potential adversarial patches in the input. Based on the effect, we present two types of defensive patches: \textit{canary} and \textit{woodpecker}.

In this paper, we propose two types of defensive patches, \textit{canary} and \textit{woodpecker}. They are used to probe or counteract potential adversarial patches, respectively. 
% In this way, potential attacks can be detected by examining the model output or the output changes after incorporating defensive patches.
In this way, potential attacks can be detected by examining or differentially analyzing the model output after incorporating defensive patches.
Furthermore, the defensive patches can naturally combine with randomized injection strategies to resist defense-aware attacks.

\textbf{Canary}, as its name suggests, is a fragile object. It is crafted to be easily perturbed by potential adversarial patches, such that it cannot be correctly recognized in an adversarial sample.
Intuitively, the canary is preferably to be placed proximal to the attacked object, where it is susceptible to potential adversarial patches.
% To make the canary work effectively, selecting an appropriate placement location is crucial.
However, determining such a position beforehand is infeasible. Fortunately,
we have found that, as a localized attack, adversarial patches cannot entirely erase all information about the target object in the model output.
In particular, while the bounding boxes of the attacked object are suppressed by the adversarial patches, making them disregarded by the model, they still maintain observable objectness or class scores.
We can identify the suppressed bounding boxes from the output, which are suspected to be related to the attacked object. This allows us to place the canary near the boxes (around or within them), \revbegin{B}{10}\hjj{making it more likely}\revend{B}{10} to be influenced by the adversarial patches.
% However, determining such a position beforehand is infeasible. Fortunately, as a localized attack, adversarial patches cannot entirely erase all information about the target object in the model output. We can identify from the output the suppressed bounding boxes suspected to be related to the attacked object. This allows us to place the canary near the boxes (around or within them), making it is more likely to be influenced by the adversarial patches.
% \hjj{The canary may appear at multiple positions within a given sample, but it is preferably to be placed in close proximity to the target object of interest that is hidden by adversarial patches, such that it is more likely to be influenced by the patches. When the target object disappears, its bounding box is probably only suppressed, but can be extracted from the output information of object detectors. We find such potentially suppressed bounding boxes and place canaries around them. Detecting the imported canary objects can tell whether there may be an adversarial patch attack. 
% As illustrated in Figure~\ref{fig-idea}(a), \st{if a potential bounding box is found in a given sample, we put a canary in its center and perform object detection.} The canary is correctly detected in the left, so the sample is considered benign; the canary is not detected in the right and thus the sample is deemed an adversarial one.}
As illustrated in Figure \ref{fig-idea}(b), we can ascertain the presence of adversarial patches in the current input by checking whether the model can detect the imported canary objects correctly.
In this example, canaries are placed to the left of the identified bounding boxes.
If any injected canary object is not correctly detected, it indicates that the input is deemed an adversarial sample.
% \xz{there may be localized patch attack that may only affect objects in a small region. As such, you canary may not work if it does not fall into the region.}

% \xz{you may want to explain intuitively why woodpecker can achieve that. The following text looks like magic}\hjj{[Updated both Woodpecker and Canary]}
% \textbf{Woodpecker} aims to impede potential adversarial patches and weaken their capability.
\textbf{Woodpecker} is designed to counteract the influence of adversarial patches on the target object, enabling it to be correctly detected once again. Essentially, adversarial patches aim to subvert the semantics of the target object to deceive object detectors. In response, woodpecker is specifically trained to carry necessary information to repair the compromised object semantics.
In a similar way, woodpecker is also placed near identified bounding boxes. 
In Figure \ref{fig-idea}(c), we can determine the malicious nature of the input by assessing whether any new objects of interest (e.g., a person) are present in the model output. Specifically, if the woodpecker recovers an object that was misidentified in the original input, the input will be classified as adversarial.

These two defensive patches can be trained and generated for an object class of interest using a limited number of benign and adversarial samples. When applying canary and woodpecker, the target model can remain unchanged. Besides, adopting them does not involve complicated online computation, and thus the overhead is limited.
% , e.g., segmentation, and thus the time overhead is limited.

\begin{figure}[t]
\centering
\includegraphics[width=\linewidth]{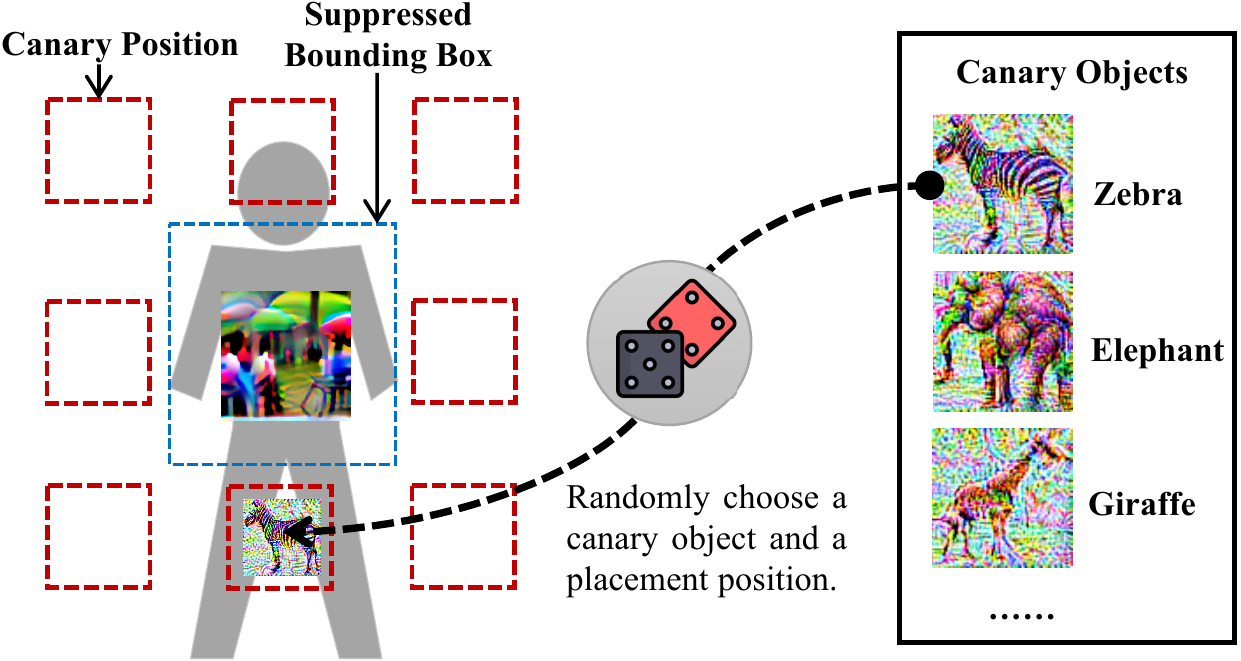}
\caption{Randomized canary patterns.}
\label{fig-random}
\end{figure}

More importantly, we also draw inspiration from the Stack Cookie ~\cite{stackcanary2000} and Address Space Layout Randomization (ASLR) ~\cite{ASLR97IEEE} techniques used in software security, and enforce a \textbf{randomization strategy} when injecting canaries and woodpeckers to enhance robustness. As an active defense measure, defensive patches can combine with various randomization patterns such as patch content and placement positions.
Taking canary as an example, as shown in Figure \ref{fig-random}, we can generate multiple canary patches containing different objects (e.g., zebra, elephant or giraffe) for each potential canary placement position. For a given input, we can randomly select a canary and place it at a random position near the identified suppressed bounding box (e.g., placing a zebra canary in the lower-middle position), or even place multiple canary patches randomly. As a result, even if the attacker fully understands all details of the canary mechanism, it would be difficult to optimize an adversarial patch that can bypass all canary patterns. Theoretically, the combination of canary objects and placement positions is infinite. In other words, defense-aware attacks will become exceedingly challenging, if not impossible.

The proposed method is evaluated with a comprehensive test set generated from three public datasets~\cite{VOC07, inria, COCO} and a physical-world one. We conducted experiments on four adversarial patch attacks ~\cite{attack_fool,attack_advtext,attack_naturalistic,attack_upc} and five object detectors ~\cite{yolov2,fasterrcnn,yolov4,yolor,yolov8}. It is worth emphasizing that we only use one attack method, i.e.,~\cite{attack_fool} or~\cite{attack_advtext}, to generate some digital-world attack samples for training canary and woodpecker. The other attacks are unknown to the obtained canary and woodpecker. 
% \xz{are any these attacks localized? You could check our ODScan paper} 
Results have shown that our method can achieve a high F1 score and is effective in defending against unknown attacks. 
% On benign samples, our method emits a low false positive rate and does not heavily decrease the model performance.
% \hjj{[TODO: normal performance!]}

Comparison with existing defense techniques has also shown that, our method can have comparable detection performance in some scenarios and achieve better performance in most scenarios. 
On benign samples, canary and woodpecker emit low false positive rates.
Meanwhile, our method also exhibits a low latency in detection, approximately 0.1 seconds per image.

We also conducted %adaptive attack 
experiments to demonstrate the effectiveness %of the proposed method 
% in defending against defense-aware attacks on 582 randomly sampled examples from four datasets. 
in defending against defense-aware attacks on 688 randomly sampled examples from four datasets. 
%%% 2024.09.20 comment out the following paragraph
% For a given example, the adaptive adversarial patch is deliberately optimized to counter defensive patches and randomized strategies, assuming the attacker has complete knowledge of them.
% % We assumed a premise highly favorable to attackers, where they know all details of the defense techniques, including canary and woodpecker objects, as well as all potential placement positions. Given an input sample, an attacker can optimize an adaptive adversarial patch to try to bypass our defense method. 
% % Two constrained randomization strategies are adopted in the experiments, i.e., six canary injection patterns (three canaries $\times$ two fixed positions) and two woodpecker patterns (one in the left or the other in the right). 
The results show that attackers cannot generate adaptive adversarial patches to completely bypass the defense even if only two constrained randomization strategies are adopted. 
% enforcing randomization in only two fixed positions using three canary or two woodpecker objects is sufficient to completely defend against all adaptive attacks (100\%). 
%%% 2024.09.20 comment out the following paragraph
% The robustness of the proposed method is demonstrated in the face of defense-aware attacks. 
% % In comparison, among 155 samples (77.5\%), the state-of-the-art defense method UDF~\cite{UDF} is evaded by adaptive adversarial patches.
% % \xz{you have a very strong cliam here, which could backfire. You may want to explai the intution why yours is so powerful}

% We believe that the proposed method is a practical technique, and can be directly used for defending adversarial patch attacks in real-world scenarios. 
Our work makes the following contributions.

\begin{itemize}
\item We present a novel and general methodology against adversarial patch attacks. Two kinds of defensive patches, \textit{canary} and \textit{woodpecker}, can be easily applied to unmodified target object detection models, and can effectively defend unknown attack methods.
\item We propose a randomization-based defensive patch deployment strategy. The robustness of canary and woodpecker is guaranteed via enforcing randomized patch injection patterns, effectively defending against defense-aware attacks.
\item
We conduct a comprehensive experiment to evaluate the proposed method, employing over ten thousand digital and physical-world samples. 
It exhibits good performance on both adversarial and benign examples.
% Our method exhibits good detection performance on adversarial example and the model performance on benign examples is only slightly decreased.
% only slight degradation of original performance on benign samples.
% \xz{you never mention if your tech degrade normal performance. How do you avoid masking existing objects?}
\end{itemize}

%% file: Background.tex
\begin{figure}[t]
\centering
\includegraphics[width=\linewidth]{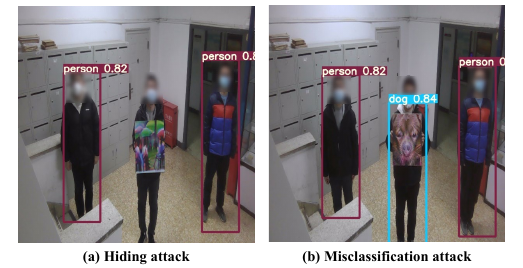}
\caption{Adversarial patch attacks examples. }
\label{fig-attack-example}
\end{figure}

\begin{figure*}[t]
\centering
\includegraphics[width=\textwidth]{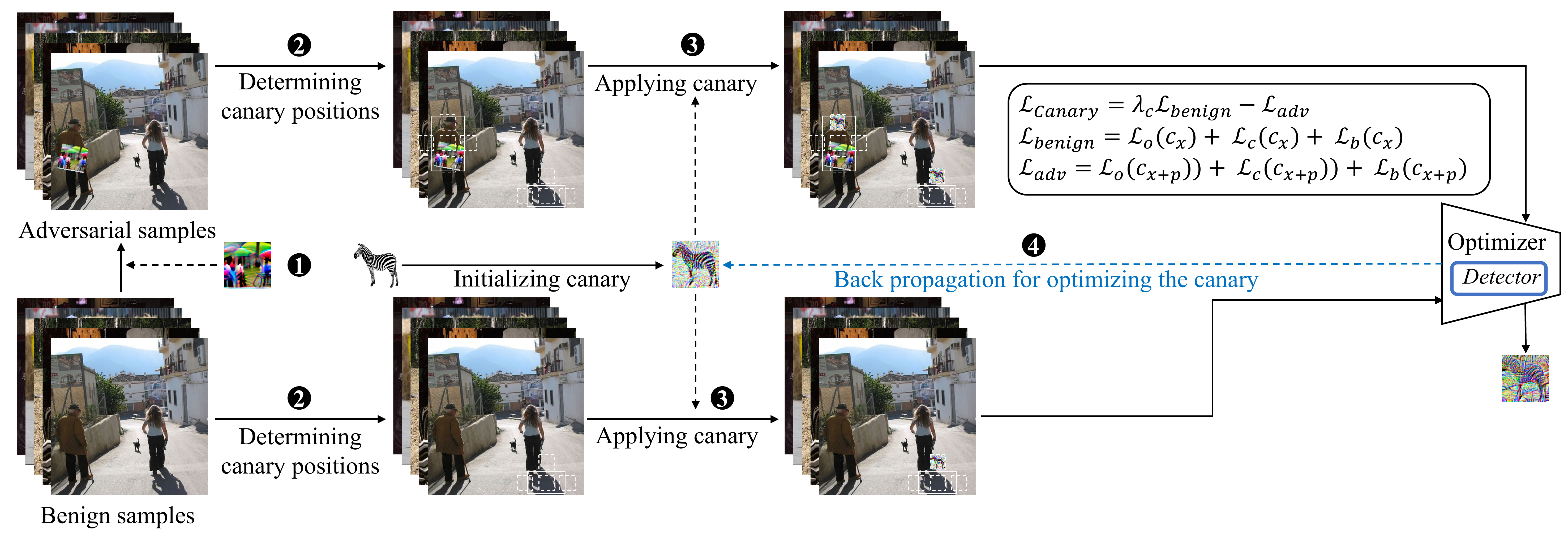}
\caption{Canary generation. \mdseries %Two sample sets are used to optimize the canary. 
It is initialized as a specific class object (e.g., a \textit{zebra}) and placed in determined positions. The canary is generated via a joint optimization with the loss $\mathcal{L}_\textit{Canary}$ in Eq.\ref{lossfun-canary}.} 
\label{fig-train-canary}
\end{figure*}

\section{Background}

\subsection{Object Detection}

The object detection task aims to identify and classify each object in an input image by predicting a set of bounding boxes. Taking YOLOv2~\cite{yolov2} as an example. A bounding box is represented as $\textit{box}=\{x $, $y$, $w$, $h$, $\textit{obj}_{\textit{score}}$, $l\}$, where $x$, $y$, $w$, $h$ represent the coordinates and dimensions, $\textit{obj}_{\textit{score}}$ represents the objectness score, and $l$ denotes the class probability distribution (object class and its score) of the object in the bounding box. The bounding boxes with low objectness will be filtered out, and then the non-maximum suppression algorithm~\cite{nms06} is used to remove redundant bounding boxes. 

Faster R-CNN ~\cite{fasterrcnn} and YOLO~\cite{yolov2,yolov4,yolor} are commonly used object detection models. Despite slight differences, the basic principles of these detectors are similar. Faster R-CNN is a two-stage object detector that first generates a set of candidate regions, called proposals, which may contain objects. These proposals are then filtered to exclude background regions without objects, and the remaining proposals undergo object classification and bounding-box regression. YOLO is a one-stage object detector that predicts object bounding boxes directly, allowing faster detection. %Both models are widely applied in physical-world applications.

\subsection{Adversarial Patch Attack}
Adversarial patch attacks \cite{attack_fool,attack_advtext,attack_naturalistic,attack_upc} aim to deceive object detectors by attaching an adversarial patch onto an object, causing it either to be missed or to be misclassified. The attacker generates an adversarial patch by attaching it to the training images and iteratively optimizing it to evade detection by the target detector or be misclassified as a specific category. Figure \ref{fig-attack-example}(a) illustrates the AdvPatch attack \cite{attack_fool}, which makes the person carrying it undetected; Figure \ref{fig-attack-example}(b) shows the UPC attack~\cite{attack_upc} that causes the person to be wrongly identified as a ``dog''.
% \xz{does the patch have to be on the object? Can it not be around the object? Do you assume it has to be on the object? Does your canary and wood pecker have to be stamped on the comrpomised object? becuase OD models have many bounding boxes . These are basic questions you should make clear in early text.}

\subsection{Threat Model}
In this study, we assume the attacker holds the initiative in the game of attack and defense. The attacker is assumed to have access to the code and data of the object detector and our defense method, understands all the algorithm and deployment details of our method, and has all pre-generated canaries and woodpeckers. In this scenario, we believe the attacker is capable of launching defense-aware attacks, generating adaptive adversarial patches against the defense mechanism to bypass it. On the other hand, we consider that the defender only has knowledge of one known attack method (without loss of generality, AdvPatch~\cite{attack_fool} or TC-EGA~\cite{attack_advtext} in this study), to generate canaries and woodpeckers, and is unaware of any other attack technique. Clearly, developing attack-specific defense mechanisms is impractical since the defender cannot guarantee prior knowledge of the attack techniques. In other words, an effective defense method must be able to withstand unknown and defense-aware attacks.

Additionally, rebuilding and redeploying a mature model is an expensive task, and we believe that modifying or retraining object detection models for defense is not always feasible.

%% file: Approach.tex
\section{Methodology}

\revbegin{A}{7}
\hjj{
Generally, given an input image $x$ and a target object detector model $O$, the defensive patches take effect as follows. We first determine the bounding boxes around target objects of interest (TOIs) that are potentially suppressed by adversarial patches from the model output $O(x)$. Then a canary $c$ or a woodpecker $w$ is placed near each identified bounding box, forming augmented input $x+c$ or $x+w$. If $c$ is not correctly detected in $O(x+c)$ or any new TOI is found in $O(x+w)$, the original input $x$ is deemed adversarial.
}
\revend{A}{7}

Generating canary and woodpecker patches is done by optimizing an initial object on a batch of adversarial and benign images. 
In this study, we generate each defensive patch for a target object class of interest, e.g., \textit{person}. %To ease the discussion, we abbreviate the target object of interest to \textit{TOI} in the following. 
% In the process of generating the defensive patches, w
We will employ different initial objects and placement positions, to prepare corresponding canary and woodpecker objects for enforcing various randomization-based deployment patterns.

\subsection{Generating Canary} 
\label{subsec:Canary}
% To utilize the canary to detect adversarial patch attacks, we first determine suitable placement positions of the canary objects in a given image, inject the canary and perform object detection on the canary-augmented image. If any canary is not detected correctly, we emit an alert, indicating a potential attack. 
% In the following describes how to generate desired canary objects. 
% As shown in Figure~\ref{fig-train-canary}, generating a canary consists of four steps, i.e., original sample preparation, position determination, training sample construction and canary optimization. 
As shown in Figure~\ref{fig-train-canary}, the workflow of generating a canary consists of four steps.

\ding{182} \textit{\textbf{Preparing adversarial and benign samples.}} 
We generate adversarial patches (e.g., from~\cite{attack_fool}) and import them to benign images. 
From the adversarial images that can attack the target object detector, we choose a few dozen as the adversarial set for training canary. 
The corresponding benign images form the benign set.

\begin{figure}[t]
\centering
\includegraphics[width=\linewidth]{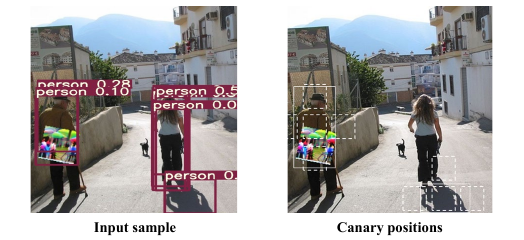}
\caption{Determining canary positions. \mdseries The undetected person objects' bounding boxes are used to get candidate boxes (white solid). Canary will be placed at the positions (white dashed) associated with the candidate boxes.}
\label{fig-position-patch}
\end{figure}

\ding{183} \textit{\textbf{Determining canary positions.}}
% \hjj{[TODO: revise; ``person'']}
% The canary patch may appear at multiple positions within the model input, but it is preferably to be placed in close proximity to the adversarial patch\xz{this should be made clear much earlier. you give people too much room to imagine how it works before this}. However, we face challenges in accurately identifying the position of adversarial patches. Our solution is to determine potential canary positions based on the underlying objects of interest that may be concealed\xz{suppressed?} by adversarial patches. 
We place canary near the bounding box that may contain a TOI but the TOI is missed by target object detector.
% % In this study, we focus on the \textit{person} object. 
% The fundamental idea is to find a bounding box that may contain an object of interest (e.g., a person), but the object is not detected.
In Faster R-CNN and early YOLO detectors, a low objectness score of a bounding box (e.g., below 0.5 in YOLOv2~\cite{yolov2}) makes the possible TOI not detected; in YOLOv8~\cite{yolov8}, the objectness is discarded and the class score is used to decide whether the TOI cannot be detected (below 0.25). 
% The scores corresponding to different detectors are called decision scores below.
% 
% In practice, even if a TOI is hidden by an adversarial patch, its objectness or class score does not disappear. 
To locate potentially suppressed bounding boxes related to TOIs, We perform an object detection on a given input, extract information from the model output and identify the
satisfactory boxes as \textit{candidate boxes}.
% and consider them as \textit{candidate boxes} that are potentially perturbed by adversarial patches. 
To minimize the risk that the candidate boxes do not contain TOIs, we require them to have an objectness or class score above a certain threshold $\tau$. 
For example, our pilot experiment in \S\ref{subsec:Eval:Parameter} determined $\tau$ = 0.05 for YOLO detectors. 
% In this study, 
We term it the \textit{candidate box threshold}. 
% \xz{the information in this chapter is important. An intuitive version of it should be moved up to the intro. Right now, a lot of questions are unanswered in intro. For example, you could explain how you find the bounding boxes that are potentially suppressed/masked by patches, and then place a canary around it. Then the readers know what you do even without details. Right now, they don't know what you do.}
% \xz{Need to address the question that the canary may cover the object}

Note that, bounding boxes may overlap. We merge overlapping boxes to a larger candidate box if the merged has a score exceeding the above threshold. In Figure~\ref{fig-position-patch}, two candidate boxes (white solid boxes) are finally identified and the left box is obtained by merging two overlapping ones.

Once candidate boxes are determined, we have the flexibility to place the canary in numerous positions within or close to the candidate box. Theoretically, the more positions available, the greater the randomness introduced. Without loss of generality, we opt to optimize canary in five positions: the center of the candidate box and an additional 30 pixels in each cardinal direction (up, down, left, and right). The determined canary positions for the sample in Figure~\ref{fig-position-patch} are indicated by the white dashed boxes. In practice, if needed, it can be easily extended to accommodate more positions. 
% \hjj{[TODO: inevitably cover object, refer to XXX -- perhaps at the end of 3.1? -- to \S 3.3?]}

\begin{figure}[t]
\centering
\includegraphics[width=\linewidth]{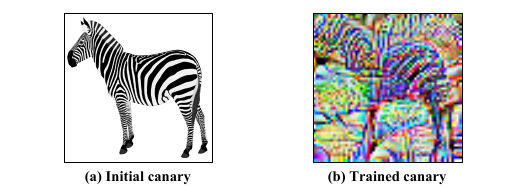}
\caption{An example of initial and a trained canary.}
\label{fig-canary-init-trained}
\end{figure}

\begin{figure*}[htb]
\centering
\includegraphics[width=\textwidth,clip,trim=0 0 0 0]{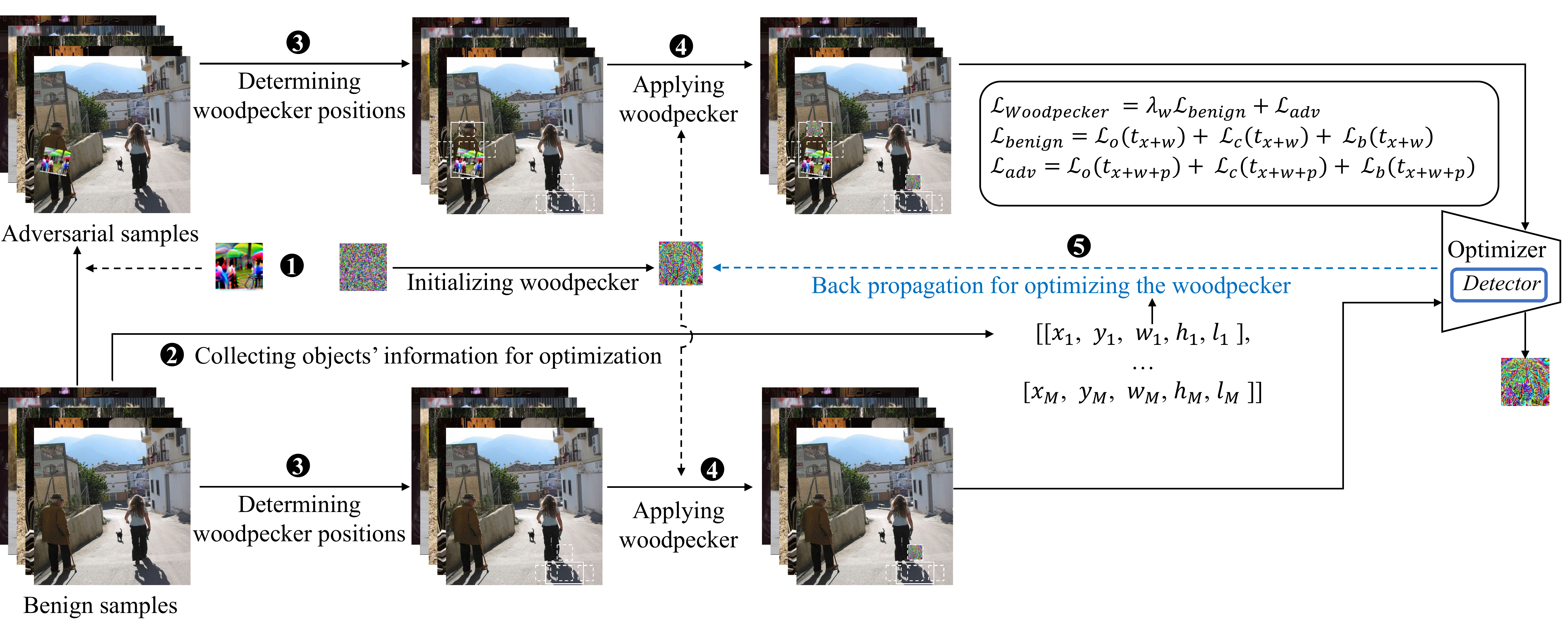}
\caption{Woodpecker generation. \mdseries %We optimize the woodpecker patch using the same sample sets for the canary. 
The woodpecker is initialized with random noises and generated via an optimization with the loss $\mathcal{L}_\textit{Woodpecker}$ in Eq.~\ref{lossfun-woodpecker}.}
\label{fig-train-woodpecker}
\end{figure*}

\ding{184} \textit{\textbf{Constructing training samples.}} 
Canary is introduced into both benign and adversarial images, making pairwise training samples.
The initial canary is an image that contains an object of a specific class, with a size determined based on a pilot experiment (e.g., 60 pixels $\times$ 60 pixels for YOLOv2 in \S\ref{subsec:Eval:Parameter}). 
Similarly, incorporating a greater variety of object classes can bring higher levels of randomness. In practice, we can choose object classes that are rarely encountered in surveillance scenes as the initial canaries. For an initial object, five canary patches can be iteratively optimized for the five previously mentioned positions.

% \lb{
Note that in a benign image, candidate boxes may be absent. We only select benign samples with candidate boxes to construct canary-augmented benign training samples.
% }
% Note that, in a benign image, we may not identify any candidate box. In this case, we choose a random position to place the initial canary, to make a canary-augmented benign sample. 
In addition, if the identified placement position extends beyond the boundary of the sample, the initial canary will be shifted to be fully included within the sample.

\ding{185} \textit{\textbf{Optimizing canary.}}
We use pairwise training samples to optimize the canary. Passing canary-augmented samples to a target object detector, we expect the detector can find all canaries in benign images but miss them in adversarial samples. 
Taking YOLOv2 as the target detector, the objective is to minimize the loss function in Eq.~\ref{lossfun-canary}.

\begin{flalign}
\label{lossfun-canary}
&\ \mathcal{L}_{\textit{Canary}} = \lambda_c\mathcal{L}_{\textit{benign}} - \mathcal{L}_{\textit{adv}} \\
&\ \mathcal{L}_{\textit{benign}} = \mathcal{L}_{o}(c_{x}) + \mathcal{L}_{c}(c_{x}) + \mathcal{L}_{b}(c_{x}) \label{lossfun-canary-benign} \\
&\ \mathcal{L}_{\textit{adv}} = \mathcal{L}_{o}(c_{x+p}) + \mathcal{L}_{c}(c_{x+p}) + \mathcal{L}_{b}(c_{x+p}) \label{lossfun-canary-adv} , 
\end{flalign}

\noindent where $\mathcal{L}_\textit{benign}$ (Eq.~\ref{lossfun-canary-benign}) corresponds to the probability loss of all canary objects being correctly detected within benign samples, and $\mathcal{L}_\textit{adv}$ (Eq.~\ref{lossfun-canary-adv}) corresponds to the loss of all canary objects being detected within adversarial samples.
% \lb{
To prevent the canary from being overly fragile and thus failing to be correctly detected in benign samples, we set a weight $\lambda_c$ greater than 1.0 for $\mathcal{L}_\textit{benign}$.
% }
The two losses consist of the objectness loss $\mathcal{L}_{o}$, class confidence loss $\mathcal{L}_{c}$ and bounding box loss $\mathcal{L}_{b}$ of canary objects.
For a canary $c$, $c_x$ denotes the object of $c$ within a benign sample $x$, and $c_{x+p}$ refers to the object of $c$ within an adversarial sample $x+p$, which is formed by incorporating an adversarial patch $p$ into $x$. 
% \newreplace{
% And $\lambda$ denotes the weight of $\mathcal{L}_{\textit{benign}}$, which is employed to prevent the canary from being excessively fragile, thereby mitigating the risk of high false positive rates.
% In this study, $\lambda$ is set to $2.0$.
% }
In YOLOv8, as the objectness is not predicted, we remove $\mathcal{L}_{o}$ from the loss function.

For an initial object, at most 50 epochs are taken to generate a desired canary. %An example is shown in Figure~\ref{fig-canary-init-trained}, with the initial zebra in Figure~\ref{fig-canary-init-trained}(a) and the trained canary in Figure~\ref{fig-canary-init-trained}(b).
% Typically, only a few rounds of iterations are required to generate a desired canary. 
For the initial zebra shown in Figure~\ref{fig-canary-init-trained}(a), we get a canary as presented in Figure~\ref{fig-canary-init-trained}(b). 
% \hjj{\st{The generated canary objects can then be applied to detect adversarial patch attacks. But it is notable that, the canary's position relative to a candidate box must be the same as in the training phase.}}

\subsection{Generating Woodpecker} \label{subsec:Woodpecker}
% % Utilizing the woodpecker to defend adversarial patch attacks takes similar steps to canary, except that the woodpecker counteracts the effect of adversarial patches and exposes the hidden person.
% Woodpecker counteracts the effect of adversarial patches and exposes the hidden TOI. 
% \xz{a possible problem is that many objects are not visible even with a benign model. Woodpeckers may cause them to appear, leading to false positives. How do you handle that?}\hjj{[Such FPs will be discussed in Exp]} 
% We take similar steps with canary to adopt woodpecker to detect adversarial patch attacks, except that we compare the detector outputs before and after introducing woodpecker. 
% If a suppressed TOI is re-detected with woodpecker, an attack is reported.

Training a woodpecker consists of five steps, as depicted in Figure~\ref{fig-train-woodpecker}. 
Three steps are similar to training a canary, including (\ding{182}) original sample preparation, (\ding{184}) position determination and (\ding{185}) training sample construction. 
Optimizing the woodpecker (\ding{186}) uses a different loss function and an extra step (\ding{183}) is taken to collect objects' information from benign samples for optimization. 
% \xz{Does it not mean that the woodpecker is only good for a certain class of objects and a particular attack? why does it work for other attacks?}

\textit{\textbf{Collecting objects' information for optimization} {\rm(\ding{183})}.}
We use the detected objects in benign images as a ground truth to guide the optimization of woodpecker. As some objects may have been hidden or misidentified due to adversarial patches in adversarial samples, we pre-collect the bounding boxes and class for detectable objects from benign samples.

\textit{\textbf{Optimizing woodpecker} {\rm(\ding{186}).}} 
Woodpecker is designed for repairing the attacked TOIs in adversarial samples. Hence the optimization objective is to make the detectable objects in benign samples to be detected in corresponding adversarial samples. Therefore, we aim to minimize the loss of objects that should be detected on both benign and adversarial samples. The relevant loss functions are shown in Eq.~\ref{lossfun-woodpecker} -- \ref{lossfun-wp-adv}. 
Note that, for YOLOv8, we exclude $\mathcal{L}_o$ from the functions.

\begin{flalign}
\label{lossfun-woodpecker}
&\ \mathcal{L}_\textit{Woodpecker} = \lambda_w\mathcal{L}_\textit{benign} + \mathcal{L}_\textit{adv} \\
&\ \mathcal{L}_\textit{benign} = \mathcal{L}_{o}(t_{x+w}) + \mathcal{L}_{c}(t_{x+w}) + \mathcal{L}_{b}(t_{x+w}) \label{lossfun-wp-benign} \\
&\ \mathcal{L}_\textit{adv} = \mathcal{L}_{o}(t_{x+p+w}) + \mathcal{L}_{c}(t_{x+p+w}) + \mathcal{L}_{b}(t_{x+p+w}) \label{lossfun-wp-adv} ,
\end{flalign}

\noindent where $t_{x+w}$ refers to a TOI % the object corresponding to a target object $t$ 
within a benign sample $x+w$, which is produced by applying a woodpecker $w$ into the original sample $x$. On the other hand, $t_{x+p+w}$ denotes the TOI %object corresponding to $t$ 
within the corresponding %an 
adversarial sample $x+p+w$ that is created by introducing adversarial patch $p$ and woodpecker $w$ in $x$. 
Similarly, we put a weight $\lambda_w$ for $\mathcal{L}_\textit{benign}$. 
The obtained information in step \ding{183} is used in Eq.~\ref{lossfun-wp-adv} to enforce the objects being detected. 

\begin{figure}[t]
\centering
\includegraphics[width=\linewidth]{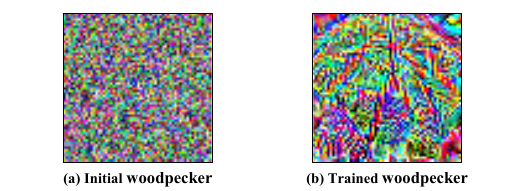}
\caption{An example initial and trained woodpecker.}
\label{fig-woodpecker-init-trained}
\end{figure}

The initial woodpecker is usually an image with random pixels, as in Figure~\ref{fig-woodpecker-init-trained}(a). 
A corresponding trained woodpecker is illustrated in Figure~\ref{fig-woodpecker-init-trained}(b). 
Similar to generating canary, we train a distinct woodpecker for each of the five placement positions.
% \hjj{\st{ and it must be applied to the same position as in the training phase}}.

\begin{figure}[t]
\centering
\includegraphics[width=\linewidth]{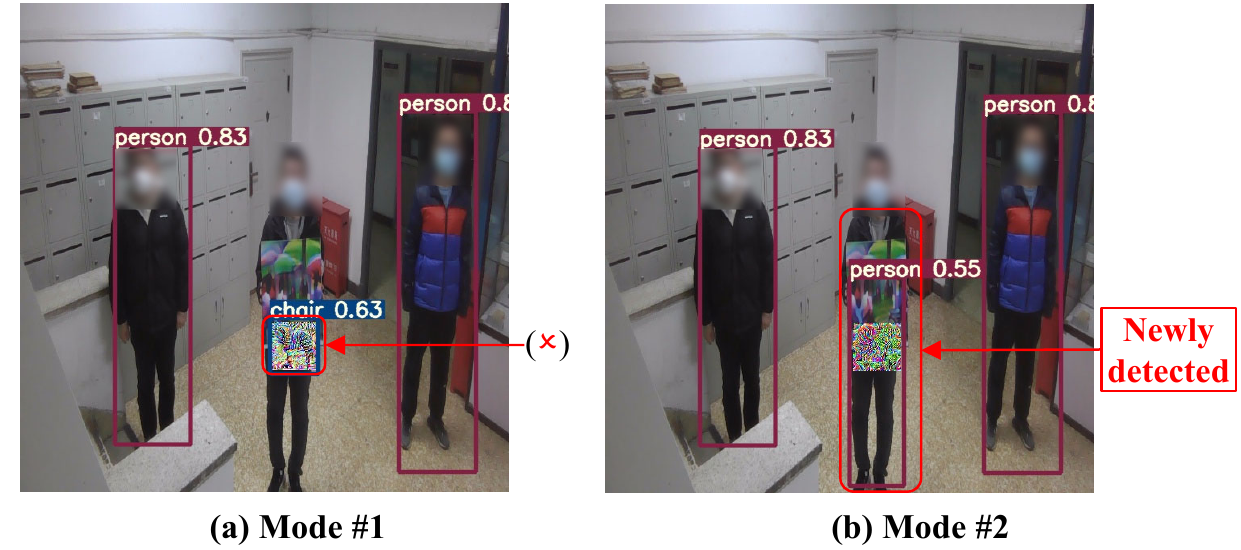}
\caption{ Examples of adopting canary alone (Mode \#1) and woodpecker alone (Mode \#2).  } 
\label{fig-cawd-detect}
\end{figure}

\subsection{Deployment and Detection} \label{subsec:Deployment}
Canary and woodpecker are two independent defensive patches that can be used individually or in combination. Accordingly, we design three deployment modes (\textbf{Mode \#1-\#3}).

\textbf{Mode \#1} utilizes canary alone.
For the current input image $x$, the positions for placing the canary are first determined, as done in canary generation. The model output $O(x)$ is leveraged to identify the candidate box within $x$. % and place a canary $c$ at the position same as in its training phase, forming a canary-augmented input $x+c$. 
\revbegin{A}{2}
\hjj{
Then a trained canary $c$ is placed at a fixed location relative to the candidate box. For example, if a canary is trained for the \textit{left} position, it will consistently be placed to the \textit{left} of the identified candidate boxes in any given sample.
}
\revend{A}{2}
Subsequently, the canary-augmented input $x+c$ is fed to the detector. %, and it is inspected whether the canary object is detected in the model output $O(x+c)$. 
If $c$ is not correctly spotted in the model output $O(x+c)$, an alert will be emitted, indicating a potential attack.
% % If $x+c$ is recognized benign, $O(x)$ is the corresponding object detection output.
Note that there may be multiple candidate boxes identified in $x$, and accordingly, multiple canaries are placed. If any single canary is missed in the output, $x$ will be considered dangerous. Conversely, if no candidate bounding boxes are identified in $x$, it is considered a benign sample and no further action is taken.
% The object detector perform only one-pass detection on the canary-augmented image. 
Figure~\ref{fig-cawd-detect}(a) shows an example of adopting canary to detect the attack in Figure~\ref{fig-attack-example}(a). 
The symbol \textcolor{red}{\ding{55}} indicates a non-detected canary and denotes an adversarial patch attack. 
% in which \ding{55} indicates a non-detected canary, \ding{51} means an identified canary and denotes a detected adversarial patch attack.

\textbf{Mode \#2} utilizes woodpecker alone. 
We take the same way to generate the woodpecker-augmented input $x+w$ for a given image $x$ and a woodpecker $w$. The model outputs, i.e., $O(x)$ and $O(x+w)$, are compared. If a TOI is detected in $O(x+w)$ but is not found in $O(x)$, we report a potential attack. 
% Both the original image and the woodpecker-augmented image need to pass the detector, to determine if an attack occurs. 
Figure \ref{fig-cawd-detect}(b) shows an example. % of adopting woodpecker to detect the attack in Figure~\ref{fig-attack-example}(a). 
The \textit{Newly detected} person in the input is explicitly marked % in  Figure \ref{fig-cawd-detect}(b) 
to indicate an attack. 

\textbf{Mode \#3} combines canary and woodpecker. 
For an input image $x$, we leverage $O(x)$ to determine the placement positions too. 
No suitable positions (i.e., no candidate boxes) imply a benign input. 
When canary $c$ can be properly imported, we generate a canary-augmented image $x+c$ and obtain the model output $O(x+c)$. If $c$ is not correctly detected, we report a potential attack and terminate the attack detection. Otherwise, we generate a woodpecker-augmented input $x+w$ and compare $O(x)$ with $O(x+w)$ to determine whether an attack may have occurred. 
% After candidate boxes are determined, an input image will pass canary and woodpecker checking in sequence with the same position information. 
% If candidate boxes are not found, or neither the canary nor the woodpecker reports an attack, the input is considered benign. Otherwise, an alert is triggered.

\textbf{Randomization}. As mentioned earlier, when deploying the canary and woodpecker, we can further raise the bar of attacks by implementing a randomization strategy, particularly for defense-aware attacks.

Specifically, for a given input $x$, we can randomly select some positions associated with candidate boxes to place one or multiple defensive patches. Moreover, there may exist multiple available defensive patches for each position, thus constituting a randomization space. 
For example, even if we only choose one out of $n$ positions to place just one canary, with $m$ distinct canaries available at each position, we can get $n \times m$ canary injection patterns. The $n \times m$ canaries, denoted as $c_1$ to $c_{m \times n}$, can be pre-generated offline. %targeting different positions with different initial objects.
Accordingly, there will be $n \times m$ potential canary-augmented inputs, i.e., $x + c_1$ to $x + c_{n \times m}$. As a result, in order to completely evade our defense, the adversarial patch must work effectively across all augmented inputs. It is very difficult, if not impossible, to generate such an adversarial patch, even for a defense-aware attacker.
% It is worth noting that expanding the randomization space is straightforward. We can just introduce more types of initial canaries, more placement positions, or simultaneously import more canaries into $x$.

Applying the randomization strategy does not introduce additional overhead, and it has been proven %to be 
necessary and highly effective against defense-aware attacks (see \S\ref{subsec:eval:adaptive}).
% Furthermore, as mentioned earlier, when deploying the canary and woodpecker, we can further raise the bar of attacks by implementing a \textbf{\textit{randomization strategy}}, particularly for defense-aware attacks.
% For an input, we can randomly select one or multiple placement positions from the identified ones and randomly choose canaries or woodpeckers from the pool of defensive patches. The available patches have been pre-generated offline, and applying the randomization strategy does not introduce additional overhead. This randomized approach has been proven to be necessary and highly effective against defense-aware attacks (see \S\ref{subsec:eval:adaptive}).

\vspace{1em}
A reasonable concern is that the imported defensive patch might cover objects in benign samples, leading to a decrease in model performance. However, apart from false positives, our method will output $O(x)$ rather than $O(x+c/w)$ for a benign sample $x$ and does not compromise the model performance. 
% Furthermore, based on our experiments, false positives are limited and lower compared to comparative methods.
Our experiments have shown that the false positives are limited.

% %%%% old canary example, replaced with {fig-cawd-detect}(a)
% \begin{figure}[t]
% \centering
% \includegraphics[width=\linewidth]{figs/example_ca_attack_res.pdf}
% \caption{ Using canary alone (Mode \#1).\hjj{[TODO: change a case where person is not detected with canary.]}}
% \label{fig-ca-detect}
% \end{figure}

% \begin{figure}[t]
% \centering
% \includegraphics[width=\linewidth]{figs/example_wd_attack_res.pdf}
% \caption{Using woodpecker alone (Mode \#2).}
% \label{fig-wd-detect}
% \end{figure}

%% file: Evaluation.tex
\section{Evaluation}
In this section, we evaluate the proposed canary and woodpecker by answering the following questions:

\begin{itemize}[leftmargin=*]
  \item \textbf{Q1}: Can our method effectively detect adversarial patch attacks and achieve comparable performance with existing methods 
  % outperform existing defense methods in detecting adversarial patch attacks 
  (\S\ref{EffectivenessAnalysis}--\S\ref{subsec:eval:non-cv})?
  \item \textbf{Q2}: Does our method significantly impact the speed of object detection (\S\ref{subsec:Efficiency})?
  \item \textbf{Q3}: Can our method effectively defend the defense-aware attacks (\S\ref{subsec:eval:adaptive})? 
\end{itemize}

\subsection{Experiment Setup} \label{subsec:Eval:Setup}

% Apart from the defense-aware attack experiments, all other experiments are conducted on a single RTX 3090 GPU with PyTorch 1.12 and CUDA 11.5.6. The defense-aware attacks are performed on a faster Tesla V100 GPU workstation.
All experiments are conducted on a single RTX 3090 GPU with PyTorch 1.12 and CUDA 11.5.6.

\textbf{Target Adversarial Patch Attacks and Object Detectors.}
In this study, we evaluate the proposed approach on detecting four types of adversarial patch attacks, namely, AdvPatch~\cite{attack_fool}, TC-EGA~\cite{attack_advtext}, Naturalistic~\cite{attack_naturalistic}, and UPC~\cite{attack_upc}. 
All the four attack methods can be directly applied to YOLOv2~\cite{yolov2}, and we have extended them to higher versions of YOLO, namely YOLOv4~\cite{yolov4}, YOLOR~\cite{yolor}, and the latest YOLOv8~\cite{yolov8}.
For another widely-used object detector, Faster R-CNN~\cite{fasterrcnn}, with the exception of AdvPatch which lacks support for attacking Faster R-CNN as mentioned in ~\cite{attack_fool}, all other three attacks were successfully adapted.
The experiments are majorly evaluated on the above five detectors and the default hyperparameter settings of the attacks were adopted to generate the adversarial patches.
If not explicitly claimed, person is the attacked object in the experiments.

\textbf{Comparative Defense Methods.}
We conducted a comparative study with \conbegin{2}\revbegin{B}{9}\hjj{five}\revend{B}{9}\conend{2} 
% We conducted a comparative study with five 
state-of-the-art defense methods (\S\ref{EffectivenessAnalysis}--\S\ref{subsec:eval:sample-specific}), namely \textit{Local Gradient Smoothing} (LGS)~\cite{LGS}, \textit{Universal Defensive Frame} (UDF)~\cite{UDF}, \textit{DetectorGuard}~\cite{DetectorGuard}, \textit{ObjectSeeker}~\cite{ObjectSeeker}, and 
% \conbegin{2}\revbegin{B}{9}
\lb{\textit{Segment and Complete} (SAC)~\cite{LiuSegDefend}}. 
% \revend{B}{9}\conend{2}
A \revbegin{B}{10}\hjj{non-computer vision (non-CV)}\revend{B}{10} defense method, PercepGuard~\cite{USENIX23_person_like_car}, is also compared (\S \ref{subsec:eval:non-cv}).

% \textbf{Comparative Defense Methods.}
% We conducted a comparative study with five state-of-the-art defense methods (\S\ref{EffectivenessAnalysis}--\S\ref{subsec:eval:sample-specific}), namely \textit{Local Gradient Smoothing} (LGS)~\cite{LGS}, \textit{Universal Defensive Frame} (UDF)~\cite{UDF}, \textit{DetectorGuard}~\cite{DetectorGuard}, \textit{ObjectSeeker}~\cite{ObjectSeeker}, and \textit{Segment and Complete} (SAC)~\cite{LiuSegDefend}. A non-computer vision (non-CV) defense method, PercepGuard~\cite{USENIX23_person_like_car}, is also compared (\S \ref{subsec:eval:non-cv}).

\input{tables/table-trainval}

\textbf{Datasets.}
We build the adversarial samples from three public digital-world datasets and a private physical-world dataset. 
The \underline{Digital-world} datasets involve three widely used datasets, i.e., \textit{VOC07} \cite{VOC07}, \textit{COCO}~\cite{COCO}, and \textit{Inria-person}~\cite{inria}. Each consists of a training set and a testing set. 
The \underline{Physical-world} dataset is gathered by ourselves. We generate printable patches and then take photos to get adversarial images (with a printed patch held by a person) and benign (without patches) samples in 14 situations (see Appendix~\ref{appendx:physicalworlddataset}).
% \revbegin{B}{2/3}
\lb{Note that we did not deliberately ensure that the training, validation and test sets had the same distribution. Additionally, we do not require the samples used to train the detector to be the same as those used for training defensive patches. We only need the training samples to contain objects of interest.
}
% \revend{B}{2/3}

\input{tables/table-parameter-threshold}

\input{tables/table-parameter-patchsize}

\subsection{Parameter Setting}\label{subsec:Eval:Parameter}
We train canary and woodpecker objects targeting Faster R-CNN/TC-EGA and YOLO/AdvPatch combinations on the VOC07 dataset. A validation set is then constructed to determine the three major parameters in this study. 
% \revbegin{C}{8}
\hjj{
It includes benign samples and their corresponding adversarial samples, randomly selected from the VOC07 dataset and generated using TC-EGA or AdvPatch.
}
% \revend{C}{8}
Table~\ref{tab-train-validation-nums} shows the number of adversarial samples in the two sets (same number for benign samples).

\textbf{Candidate Box Threshold.} 
To determine an appropriate $\tau$ used in \S\ref{subsec:Canary}, we conducted a pilot study using the validation set for each of the five object detectors.
We determine an appropriate $\tau$ based on 
% by summing up the quantities of samples that satisfy 
the following three conditions. 
(1) There exists a candidate box around an attacked person. 
(2) There are no candidate boxes around a non-attacked person. 
(3) For an attacked person, the Intersection over Union (IoU) \cite{IoU} between the identified candidate box and the corresponding true box in the benign sample exceeds 0.5. 
For target detectors, the value greater than 0.5 indicates that the two boxes mark the same object. 
% \revbegin{C}{7}
\hjj{Each validation sample is then scored based on how many of the conditions it meets and $\tau$ is selected with the highest scores.}
% \revend{C}{7}
% The highest scoring $\tau$ value is selected as the threshold. 
As shown in Table \ref{tab:threshold}, $\tau$ is determined as 0.075 for Faster R-CNN and 0.05 for YOLO detectors.

\textbf{Defensive Patch Size.} 
% We determine a proper size of defensive patches by evaluating the effectiveness using the different sizes of defensive patches. We conduct experiments on each object detector individually. 
To determine a proper size of defensive patches, we place five trained canaries at five determined positions, respectively (refer to Figure~\ref{fig-position-patch}), making five variants for each sample. Each position is placed with a distinct canary (i.e., \textit{zebra}, \textit{cow}, \textit{toaster}, \textit{boat}, and \textit{elephant}).
Similarly, we respectively place five distinct woodpecker objects at five positions in each sample.
% \revbegin{A}{3}
\lb{For each target detector, we will explore patch sizes ranging from 20×20 to 200×200 with a 20-pixel step size.
}
% \revend{A}{3}
Averaged F1 score for each scenario on the validation set is computed.

As presented in Table~\ref{tab-patch-size}, the canary size is set as 120$\times$120, 60$\times$60, 60$\times$60, 120$\times$120 and 80$\times$80 for Faster R-CNN, YOLOv2, YOLOv4, YOLOR and YOLOv8, respectively. 
% The woodpecker size is determined as 120$\times$120, 60$\times$60, 60$\times$60, 140$\times$140 and 80$\times$80 for the respective detectors.
The woodpecker size is the same as the canary size, except that for YOLOR, it is 140$\times$140.

\textbf{Weights for $\mathcal{L}_\textit{benign}$.} In this study, we have adopted a balanced strategy to determine the weights $\lambda_c$ and $\lambda_w$ used in Eq.~\ref{lossfun-canary} and Eq.~\ref{lossfun-woodpecker}, respectively. We aim for the resulting canary and woodpecker to have good detection performance while also not causing too many false positives. Using a method similar to the above, $\lambda_c$ is set to 2.0 and $\lambda_w$ is 1.0.

\input{tables/table-compare-all-new0918}

\subsection{Effectiveness} \label{EffectivenessAnalysis}
\textbf{Preparation.}
We systematically generated adversarial patches for 19 combinations of attack methods and target object detectors, using the training set from the Inria dataset.
% These patches were then strategically placed within the images in the physical-world dataset and the testing set of the public datasets.
These patches were then strategically integrated into the testing samples of public datasets or printed for the purpose of creating physical adversarial examples.
To obtain as many adversarial samples as possible, the positions, the sizes and the angles of the patches are adjusted carefully. 
Eventually, we dedicated hundreds of hours to obtain a total of 6,800 digital-world and 1,960 physical-world adversarial samples, with their 6,071 distinct original counterparts serving as benign samples.

We take the same training set as in \S\ref{subsec:Eval:Parameter} to generate the defensive patches and deploy three defense modes (see \S\ref{subsec:Deployment}) to detect %(known and unknown) 
adversarial patch attacks on five object detectors. 
In theory, a random defensive patch can be placed at a random position. 
Without loss of generality, we place a fixed defensive patch (\textit{zebra}) at the determined center position in the experiment.
In this study, we have chosen to utilize F1 Score as the evaluation metric, as it considers both precision and recall, offering a more comprehensive assessment of the model performance.

\input{tables/table-false-positive}

\input{tables/table-target-car}

\input{tables/table-sample-specific}

\textbf{Results.}
In Table~\ref{tab-compare-all}, the AdvPatch-related and TC-EGA/Faster R-CNN-related rows (a total of 7 rows) present the performance of \textbf{detecting known attacks} (where the attack methods are known to canary and woodpecker generation). 
% Our method demonstrates better performance than existing defense methods.
% Specifically, Mode \#1 (only canary) exhibits higher F1 scores than the compared methods in 
% 79\% of scenarios (15 out of 19). Mode \#2 (only woodpecker) outperforms them in 73\% of scenarios (14 out of 19).
% Even in cases where our method is slightly inferior, both Mode \#1 and Mode \#2 demonstrate comparable performance. Mode \#3 (combining canary and woodpecker) outperforms the comparison methods in all scenarios.
% \conbegin{2}\revbegin{B}{9}
\lb{To ensure a fair comparison, we retrained SAC using adversarial patches generated by AdvPatch and TC-EGA. Overall, our method outperforms the five compared methods across all three deployment modes: Mode \#1 (canary only), Mode \#2 (woodpecker only), and Mode \#3 (combining canary and woodpecker).}

\lb{Notably, SAC outperformed our method in two specific scenarios (AdvPatch/YOLOv2 and AdvPatch/YOLOv8), even in Mode \#3. This performance advantage stems from SAC's ability to train a segmenter that identifies the adversarial patches, effectively masking the corresponding pixels. When faced with previously seen adversarial patches, the segmenter may leverage the learned features, leading to strong detection performance. However, SAC shows poor generalization when confronted with unknown attacks. Furthermore, as described in \S\ref{subsec:eval:sample-specific}, for unseen adversarial patches generated from known attacks (AdvPatch), SAC also performs poorly.}
% \revend{B}{9}\conend{2}

% The other 52 rows (with a light gray background) in Table~\ref{tab-compare-all} show the performance of \textbf{detecting unknown attacks} (The attack methods are unknown to canary and woodpecker generation). 
% In fact, either canary or woodpecker is trained targeting one kind of adversarial attack (see \S\ref{subsec:Eval:Setup}) and we directly apply the defensive patch to detect the other kinds of attacks. 
% In this situation, our proposed method still works well, with Mode \#1 and Mode \#2 outperforming the compared methods in 79\% (41 out of 52) and 67\% (35 out of 52) of scenarios, respectively. Mode \#3 consistently outperforms them in all scenarios.

Table~\ref{tab-compare-all} also shows the performance of \textbf{detecting unknown attacks} (where the attack methods are unknown to canary and woodpecker generation). 
In fact, either canary or woodpecker is trained targeting one kind of adversarial attack (see \S\ref{subsec:Eval:Setup}) and we directly apply the defensive patch to detect the other kinds of attacks. 
In this situation, our proposed method still works well. 
\lb{All three deployment modes outperform every compared method.}
% \revbegin{C}{10}
\hjj{We intuitively think that although different attack methods generate adversarial patches in various ways, the obtained patches might perturb the target object in similar ways, allowing defensive patches to exhibit transferability.}
% \revend{C}{10}

% It is worth noting that among the four comparative methods, ObjectSeeker exhibits the best performance. However, it is unfortunate that its speed is slow (see \S\ref{subsec:Efficiency}).

On the whole, across the entire testing set, all three modes of our method outperform the compared methods in terms F1 scores, as indicated by the last row in Table~\ref{tab-compare-all}. 
% Appendix \ref{appendx:casestudy} shows a few case studies. 

% \revbegin{D}{2}
\hjj{We further inspected the false positives (FPs).
The results are shown in Table~\ref{tab-false-positive}. 
Canary and woodpecker incorrectly report 4.9\% and 1.4\% of benign images as adversarial, respectively. Both are better than LGS (5.8\%), UDF (5.3\%), DetectorGuard (15.5\%) and ObjectSeeker (28.2\%).}
\lb{SAC exhibited a very low false positive rate; however, its F1 score is only 0.445.}
% \revend{D}{2}

% Mode \#1 and Mode \#3 incorrectly report \st{7.4\%} \newreplace{4.9\%} and \st{7.9\%} \newreplace{5.5\%} of benign images as adversarial, respectively, \st{slightly worse than}\newreplace{comparable with} LGS (5.8\%) and UDF (5.3\%) but much better than DetectorGuard (15.5\%) and ObjectSeeker (28.2\%). 
% But notably, %the two modes detect more adversarial attacks as shown in the R columns in Table~\ref{tab-compare-all}. 
% Mode \#2 has a rate (1.4\%) much lower than all the other methods. 

% We have also tested the impact of different initial classes and placement positions, and comparable performance is observed. 
% The detailed data can be found in Table~\ref{tab-dfpatch-class-position} of Appendix \ref{appendx:initialplacement}.

% \revbegin{B}{7}
\hjj{
We have also investigated the impact of different initial classes and placement positions. 
The detection performance is not largely affected by the categories or positions. 
The results can be found in Table~\ref{tab-dfpatch-class-position} of Appendix \ref{appendx:initialplacement}.
}

\input{tables/table-efficiency}

\begin{figure}[t]
    \centering
    \includegraphics[width=\linewidth]{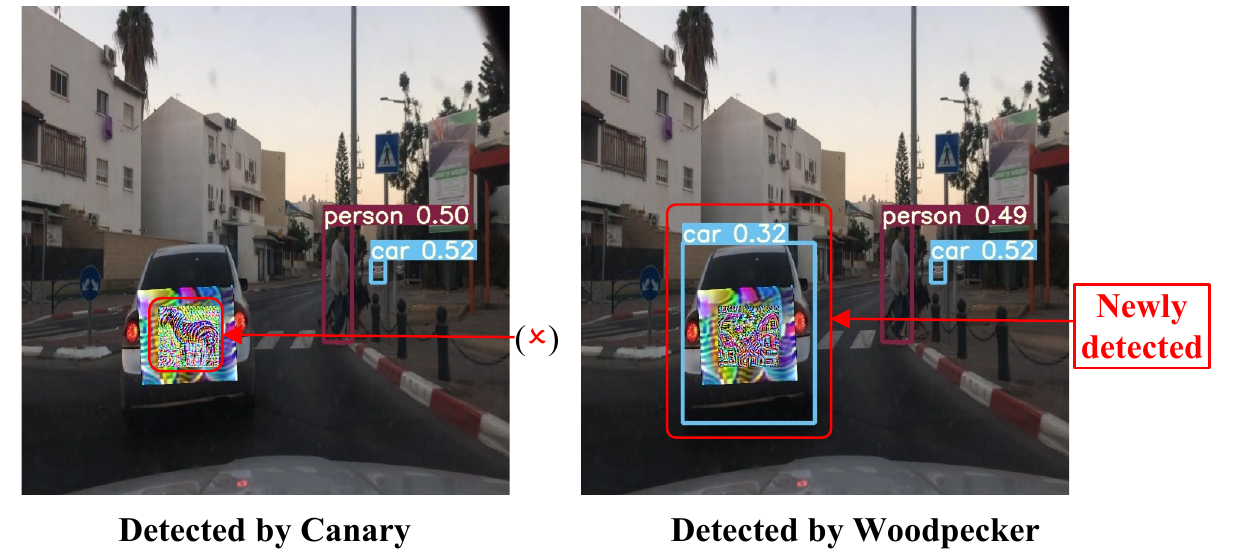}
    \caption{Detecting the adversarial attacks targeting cars.}
    \label{car_adv_cawd}
\end{figure}

\subsection{Detecting Attacks Targeting Different Types of Objects}
\label{subsec:eval:diff-types}
% The proposed method is not limited to defending the attacks targeting the person object. It can protect other object types as well. We take the car object, which is also a common target of attacks~\cite{Suryantocvpr22_target_car, Duanijcai22_target_car, attack_upc}, as an example to illustrate the versatility of our method.
% % \lb{
% We employed a newer version of YOLO (i.e., YOLOv8), as the target model, and employed AdvPatch to generate 100 adversarial examples with hidden car objects on the BDD100K dataset~\cite{yu2020bdd100k}.
% % }
% % For YOLOv8, we employed AdvPatch to generate 100 adversarial examples with hidden car objects on the BDD100K dataset~\cite{yu2020bdd100k}
% As illustrated in Figure \ref{car_adv_cawd}, the adversarial samples can be effectively detected with our defensive patches.
% The experimental result is presented in Table~\ref{tab-target-car}. The proposed method outperforms the four comparative methods, achieving the highest F1 score. It is demonstrated that our method can be applicable to other types of protected objects.

\conbegin{1}
\hjj{
The proposed method is not limited to defending the attacks targeting the person object. It can protect other object types as well. We take four other types of objects, i.e., \textit{car}, \textit{bicycle}, \textit{dog} and \textit{train}, as an example to illustrate the versatility of our method.
We employed a newer version of YOLO (i.e., YOLOv8), as the target model, and employed AdvPatch to generate as many adversarial samples as possible.
The experimental results are presented in Table~\ref{tab-target-car}. Both Mode \#1 and Mode \#3 outperform the five comparative methods in all scenarios, achieving the highest F1 scores. 
Mode \#2 does not perform the best for dog and train, but achieves better overall performance than the other five methods. It is demonstrated that our method can be applicable to other types of protected objects.
Figure~\ref{car_adv_cawd} shows an example, in which the adversarial samples target the car object can be effectively detected with our defensive patches.}
\conend{1}
% % For YOLOv8, we employed AdvPatch to generate 100 adversarial examples with hidden car objects on the BDD100K dataset~\cite{yu2020bdd100k}
% As illustrated in Figure \ref{car_adv_cawd}, the adversarial samples can be effectively detected with our defensive patches.
% The experimental result is presented in Table~\ref{tab-target-car}. The proposed method outperforms the four comparative methods, achieving the highest F1 score. It is demonstrated that our method can be applicable to other types of protected objects.

\subsection{Detecting Sample-Specific Attacks}
\label{subsec:eval:sample-specific}

Attackers can create adversarial patches specifically designed for individual samples. We evaluated the ability of our method to detect such sample-specific attacks. Note that generating sample-specific adversarial patches can be time-consuming. Without loss of generality, we randomly selected 25 samples from each of the four datasets and employed AdvPatch to generate 100 sample-specific adversarial examples targeting %the latest 
YOLOv8.

Surprisingly, the experiments have demonstrated that our method performs even better in detecting sample-specific attacks compared to universal ones. As shown in Table~\ref{tab-sample-specific}, the proposed method achieves perfect F1 scores (1.0) on most datasets, outperforming the \conbegin{2}\revbegin{B}{9}\lb{five}\revend{B}{9}\conend{2} comparative methods. It is worth mentioning that we did not generate sample-specific canaries and woodpeckers during the experiments; instead, we directly employed the defensive patches generated for universal attacks (\S\ref{EffectivenessAnalysis}). This indicates that our method can effectively counter sample-specific attacks without the need for any adjustments.

\subsection{Comparison with Non-CV Defense Method}
\label{subsec:eval:non-cv}
Apart from employing various CV techniques for defending adversarial attacks, Man et al.~\cite{ USENIX23_person_like_car} introduced a novel non-CV approach called PercepGuard. 
As a SOTA method, PercepGuard leverages the spatiotemporal properties of the detected object and constructs a RNN-based sequence classifier. The classifier is used to identify the class of the target object based on its bounding box sequence, e.g., determining whether it is a car or a pedestrian. PercepGuard can detect misclassification attacks targeting autonomous systems by cross-checking the consistency between the object track and class predictions.

We conducted a comparative experiment with PercepGuard. Regrettably,
% its dataset is not publicly available, and 
generating test object sequences is not a trivial task. We made every effort to construct 147 adversarial object sequences targeting YOLOv3 on the BDD100K dataset, intentionally causing misclassification of cars as pedestrians, following the settings of PercepGuard. The experimental results show that the three modes (Mode \#1$\sim$\#3) of our method achieved F1 scores of 
0.812, 0.831, and 0.834, 
respectively, all comparable to PercepGuard (0.781).

\input{tables/table-adaptive-attack}

\subsection{Efficiency} \label{subsec:Efficiency}
Efficiency is a crucial consideration for the practical application of defenses. We present the average run time of our method with other comparative methods in Table~\ref{tab-runtime}. 
% All experiments are carried out on the VOC07 dataset. 
Most of the detection can be finished within 0.1 seconds on average, except that DetectorGuard takes more than 0.1 seconds and ObjectSeeker takes more than 2.2 seconds for adversarial images and 2.0 seconds for benign ones. 
Our method shows higher time cost than LGS, UDF and SAC in some cases. 
However, considering that our method has a better detection performance (see \S\ref{EffectivenessAnalysis}) and the detection costs only about 0.1 seconds or less, the time overhead is limited. 
In fact, our method can sustain a video frame rate of about ten fps. It is acceptable in real-world scenarios of detecting adversarial patch attacks.

It is worth noting that the combination of canary and woodpecker does not result in a doubling of the time cost.
The causes can be three fold. 
First, positioning the defensive patches is done only once for an input image; second, many benign samples do not have any candidate boxes, making canary and woodpecker checks skipped; and third, most adversarial samples will be reported by the canary check and only a few will be passed to the woodpecker check.

% \begin{figure*}[t]
%     \centering
%     \includegraphics[width=\textwidth]{figs/compare_block_ca.pdf}
%     \caption{Comparisons between canary (a) and other types of content blocks (c--f). \mdseries 
%     The results clearly prove that detecting attacks using canary relies on the interference of adversarial patches rather than by covering the attack patch.  
%     }
%     \label{fig-compare-ca}
% \end{figure*}

\subsection{Detecting Defense-aware Attacks} 
\label{subsec:eval:adaptive}
Attackers may launch an adaptive attack when they possess knowledge of the defense. To further evaluate the proposed method, we employ an enhanced AdvPatch technique to train adaptive adversarial patches. Expectation Over Transformation (EOT)~\cite{athalye2018synthesizing} is leveraged in the patch training. EOT has been proven to be effective in generating robust adversarial samples against various defense techniques~\cite{DBLP:conf/nips/TramerCBM20}, and has been partially supported in AdvPatch. We implement the unsupported EOT features (e.g., \textit{translation}) to generate stronger patches. We apply the unchanged defense method to detect the attacks for demonstrating the effectiveness of our method in detecting such defense-aware attacks.

% fjn-0927: change ulmost to utmost ?
Note that, the cost is high to generate adaptive attack patch for each input image targeting each detector. We concentrate only on YOLOv8 and try our utmost to generate adaptive adversarial patches for 
% \revbegin{A}{4}\hjj{382 images, as shown in the \#Adv column in Table~\ref{tab-adaptive-attack}.}\revend{A}{4}
488 images, as shown in the \#Adv column in Table~\ref{tab-adaptive-attack}.
Implementing AdvPatch on YOLOv8 takes Eq.~\ref{eq:loss:ori advpatch_v8} as the loss function.

\begin{equation}
\label{eq:loss:ori advpatch_v8}
\mathcal{L}_\textit{AdvPatch} = \alpha \mathcal{L}_\textit{nps}+\beta \mathcal{L}_\textit{tv} + \mathcal{L}_\textit{cls},
\end{equation}

\noindent where $\alpha$ and $\beta$ denote the weights specifically assigned in AdvPatch, $\mathcal{L}_\textit{nps}$ represents the non-printability score of the patch, $\mathcal{L}_\textit{tv}$ denotes its total color variation, and $\mathcal{L}_\textit{cls}$ is the maximum class score of the attacked class in the image.
% \lb{
% $\mathcal{L}_\textit{nps}$ and $\mathcal{L}_\textit{tv}$ ensure the adversarial patch is printable and has smooth color transitions, while 
Among them, $\mathcal{L}_\textit{cls}$ is used for hiding the target object.
% }

% \lb{
An ideal adaptive adversarial patch $apc$ targeting canaries should ensure that the introduced canary can be detected in the presence of $apc$, while also being able to hide the target person.
% }
% An ideal adaptive adversarial patch $apc$ targeting canaries should be capable of hiding the target person while ensuring the canary is still detectable.
We use the loss function $\mathcal{L}_\textit{ac}$ shown in Eq.~\ref{eq:loss:adaptive canary_v8_new} to generate such adversarial patches. In $\mathcal{L}_\textit{ac}$, $\mathbb{C}$ is the set of canaries equipped in the defense.
For a given canary $c$, $c_{x+apc}$ refers to the object of $c$ within the sample $x+apc$, which is formed by incorporating $apc$ into the image $x$.
% \lb{
$\mathcal{L}_\textit{adv}(\cdot)$ is the loss of $c$ being detected (presented in Eq.~\ref{lossfun-canary-adv}), which is introduced to enable the detection of canary $c$ in samples where the adversarial patch $apc$ is present.
% }
% $\mathcal{L}_\textit{adv}(\cdot)$ is the loss of $c$ being detected (presented in Eq.~\ref{lossfun-canary-adv}).

% \begin{equation}
% \label{eq:loss:adaptive canary_v8}
% \mathcal{L}_\textit{ac} = \mathcal{L}_\textit{AdvPatch} + \underset{c\in \mathbb{C}}{\sum}\mathcal{L}_\textit{adv}(c_{x+apc})
% \end{equation}

% \newreplace{
\begin{equation}
\label{eq:loss:adaptive canary_v8_new}
\mathcal{L}_\textit{ac} = \underset{c\in \mathbb{C}}{\sum} \left[ \mathcal{L}_\textit{AdvPatch} + \mathcal{L}_\textit{adv}(c_{x+apc}) \right]
\end{equation}
% }

Similarly, an adaptive adversarial patch $apw$ targeting woodpeckers should be able to hide the target person even in the presence of woodpeckers, which is generated with the loss function $\mathcal{L}_\textit{aw}$ in Eq.~\ref{eq:loss:adaptive woodpecker_v8}. 
In $\mathcal{L}_\textit{aw}$, $\mathbb{W}$ is the set of equipped woodpeckers, and $x+apw+w$ refers to the input that is created by adding $apw$ and a woodpecker $w$ in the image $x$.
% \lb{
Namely, $apw$ is trained using the woodpecker-augmented inputs rather than the original one.
% }

\begin{equation}
\label{eq:loss:adaptive woodpecker_v8}
\mathcal{L}_\textit{aw} = \underset{w\in \mathbb{W}}{\sum} \left[ \alpha \mathcal{L}_\textit{nps}+\beta \mathcal{L}_\textit{tv} + \mathcal{L}_\textit{cls}({x+apw+w}) \right]
\end{equation}

Given the adaptive attack patches, though we can randomly place the defensive patches, here we evaluate the effectiveness of our method in one case that could provide very limited capability of randomization, i.e., the placement of the defensive patches is fixed to two positions. 
We propose four deployment strategies (\textbf{Dep \#1} $\thicksim$ \textbf{\#4}).
\textbf{Dep \#1} deploys a \textit{giraffe} canary at the left side of the candidate box.
\textbf{Dep \#2} randomly places a canary at the left or right sides. There are three fixed choices of canary objects, i.e., \textit{giraffe}, \textit{elephant} and \textit{zebra}. 
\textbf{Dep \#3} deploys a woodpecker at the left side.
And \textbf{Dep \#4} randomly places a woodpecker at the left or right side.

We choose UDF~\cite{UDF}, which shows a fast speed and good detection performance in previous experiments, as the comparison baseline. We follow its paper to train a shielding frame $s$. Similarly to the generation of $apw$, we use the loss $\mathcal{L}_\textit{au}$ in Eq.~\ref{eq:loss:adaptive UDF_v8}  to train the adaptive adversarial patch $apu$ targeting UDF.

\begin{equation}
\label{eq:loss:adaptive UDF_v8}
\begin{split}
\mathcal{L}_\textit{au} = \alpha \mathcal{L}_\textit{nps}+\beta \mathcal{L}_\textit{tv}
 + \mathcal{L}_\textit{cls}(x+apu+s)
\end{split}
\end{equation}

The result is shown in Table~\ref{tab-adaptive-attack}.
% Facing a single defensive patch, either a canary (Dep \#1) or a woodpecker (Dep \#3), the adaptive attack can bypass the detection in \revbegin{A}{4}\hjj{275 and 199}\revend{A}{4} samples, respectively.
Facing a single defensive patch, either a canary (Dep \#1) or a woodpecker (Dep \#3), the adaptive attack can bypass the detection in 338 and 228 samples, respectively.
It illustrates the high risk of this kind of deployment. 
However, none of the samples protected by Dep \#2 or Dep \#4 can be bypassed, even though their patterns and positions are very limited.
% Furthermore, from Table~\ref{tab-adaptive-attack}, we can also see that generating the patches for Dep \#2 and Dep \#4 consumes more than \revbegin{A}{4}\hjj{29 and 14}\revend{A}{4} days, respectively.
Furthermore, from Table~\ref{tab-adaptive-attack}, we can also see that generating the patches for Dep \#2 and Dep \#4 consumes more than 37 and 18 days, respectively.
However, even after spending so much time, we were unable to successfully generate an effective adaptive adversarial patch. The results strongly demonstrate the robustness of the proposed method enhanced with the randomization strategies.

% We observe that \revbegin{A}{4}\hjj{286 (74.9\%)}\revend{A}{4} adaptive samples can compromise UDF.
We observe that 357 (73.2\%) adaptive samples can compromise UDF.
The poor robustness might result from the fixed deployment pattern, as in Dep \#1 and \#3. 
Even if the shielding frame imposes a powerful disturbance, an adaptive attack can learn to cope with the perturbations.

% \revbegin{A}{5}
\lb{
In addition, we employ the unseen attack method Naturalistic to conduct adaptive attack experiments, which has not been used in training defensive patches. Similarly, we modify its loss function to generate adaptive adversarial patches. We spent approximately two weeks generating 200 adversarial samples across the four datasets, 50 for each datasets. The experimental results demonstrate that our method remains robust against unseen adaptive defense-aware attacks, with none of the attack samples successfully bypassing either Dep \#2 or Dep \#4.
}
% \revend{A}{5}

\subsection{Hiding Candidate Boxes} 
\label{sec:Eval:robustness}

In our method, we need to identify candidate bounding boxes for placing defensive patches. One potential evasion is to hide the bounding boxes related to the hidden object as well. This can be achieved by using the adversarial patch to reduce the objectness score or class score of the boxes below the threshold $\tau$ (0.05). However, we found that doing so would result in a significant increase in the size of the adversarial patch, heavily compromising its utility.

\begin{figure}[t]
  \centering
  \includegraphics[width=\linewidth]{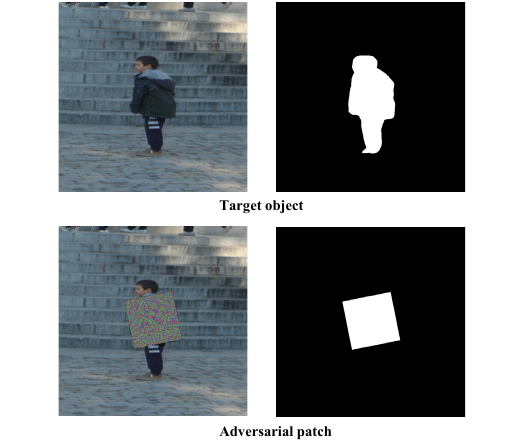}
  \caption{An adversarial patch to conceal candidate boxes.}
  \label{fig-iou-example}
\end{figure}

\begin{figure*}[t]
    \centering
    \includegraphics[width=\textwidth]{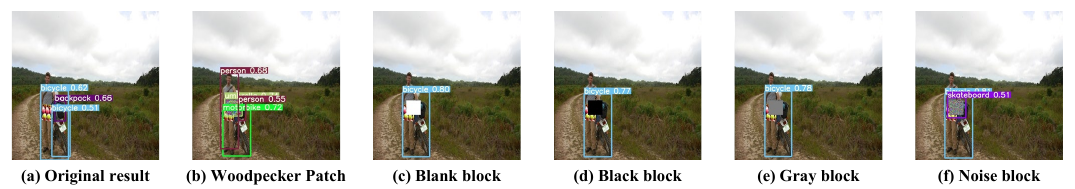}
    \caption{Comparisons between woodpecker (b) and other types of content blocks. \mdseries
    Woodpecker can make victim objects re-detected  while the other kinds of content are incapable to recover the attacked objects.
    }
    \label{fig-compare-wd}
\end{figure*}

\begin{figure}[!t]
\centering
\includegraphics[width=\linewidth]{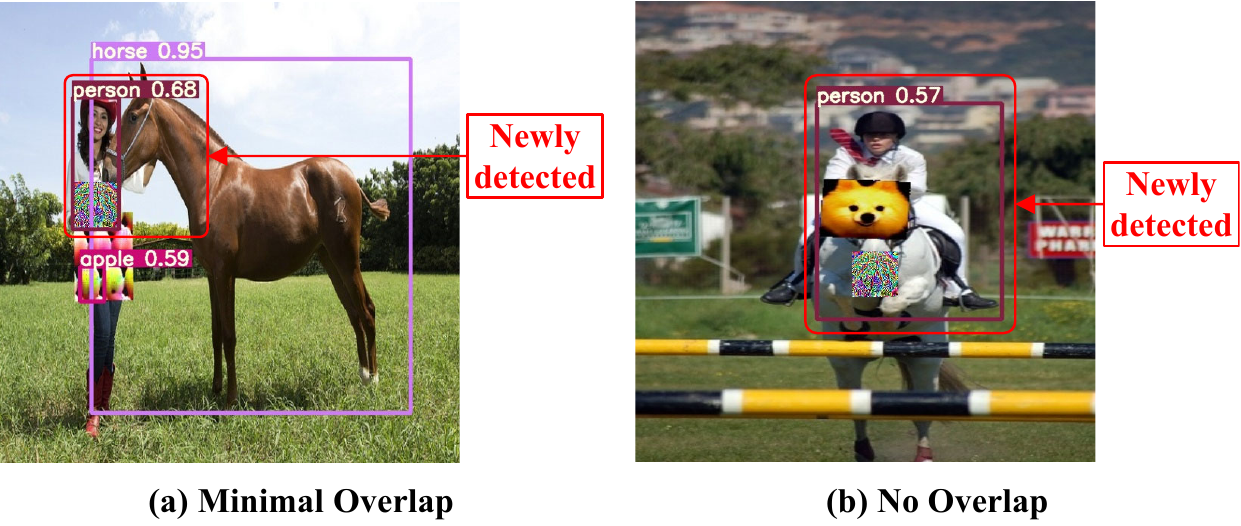}
\caption{
    % \lb{
    Woodpecker recovers the attacked object even in cases with minimal or no overlap with the adversarial patch.
    % }
    % Effectiveness of woodpecker with little overlap of adversarial patches. \mdseries In (a), woodpecker covers a very small portion, and in (b), it does not overlap the attack patch. However, the woodpecker effectively recover the victim objects.
}
\label{fig-occlusion-wd}
\end{figure}

Specifically, we use a modified AdvPatch to generate adversarial patches targeting YOLOv8, with the optimization objective of minimizing the class score of the target bounding box. 
% \revbegin{C}{3}
\hjj{The loss function was adapted as follows to not only hide the target object but also to reduce its class score to below $\tau$:}

\hjj{
\begin{equation}
L = \alpha L_{nps} + \beta L_{tv} + \lambda[\text{max(class score)} - \tau]
\end{equation} 
}

\hjj{The initial size of the adversarial patch was consistent with the settings in AdvPatch. Optimization was terminated if the learning rate fell below 1e-7 or if the iteration count reached 10,000. 
% \revbegin{C}{4}
If the generated adversarial patch did not meet the requirements, the patch size was increased by 10\% and optimization was restarted.
}
% \revend{C}{3/4}

% \revbegin{A}{6}\revbegin{C}{6}\revbegin{D}{1}
Results from \hjj{40} randomly selected samples revealed that significantly enlarging the adversarial patch is necessary to reduce the class score below the threshold. 
\hjj{Compared to patches without concealed bounding boxes, the average patch size increases from 11,275 to 53,913 pixels and the ratio to the target object area increases from 0.218 to 1.050 (3.82 times larger).}
Obviously, such large patches are impractical in real-world scenarios. 
% \revend{D}{1}\revend{C}{6}\revend{A}{6}

For example, for the child object in Figure \ref{fig-iou-example}, with an area of 33,617 pixels, the class score of its bounding box decreases below the threshold only when the adversarial patch is expanded to 27,198 pixels.
At this point, the ratio of their areas is 0.809\hjj{, nearly the same size of the target object}. %, and the IoU between them is 0.589.
In other words, the adversarial patch needs to cover a significant portion of the target object to effectively conceal the associated bounding boxes.
Using such a large adversarial patch would be too conspicuous and impractical for real-world attacks. 
% \revbegin{C}{5}
% \hjj{
In contrast, the patches in Figures~\ref{fig-cawd-detect} and~\ref{car_adv_cawd} are smaller, with the area ratio of the patch to the target object of 0.425 and 0.482, respectively. They cover a small portion of the target object, resulting in better integration and increased practicality in real-world applications.
% }
% \revend{C}{5}

In summary, concealing candidate boxes while ensuring the utility of adversarial patches is extremely challenging. Our method exhibits considerable robustness in this aspect.

% \textcolor{red}{\bf [C-Q6: In the revised paper, we plan to provide a more comprehensive analysis of size distribution on more adaptive attack samples. Relevant examples will be included in the appendix of the paper.]}

%% file: tables/table-trainval.tex
\begin{table}[t]
\centering
\caption{Numbers of adversarial samples in the training and validation sets.}
\begin{tabular}{cccc}
\hline
% Detector & Attack Method & \makecell{\#Training samples} & \makecell{\#Validation samples} \\ \hline
Detector & Attack Method & \makecell{\#Training} & \makecell{\#Validation} \\ \hline

Faster R-CNN & TC-EGA & 16 & 50 \\ 
YOLOv2 & AdvPatch & 120 & 100 \\
YOLOv4 & AdvPatch & 60 & 50 \\
YOLOR & AdvPatch & 120 & 100 \\
YOLOv8 & AdvPatch & 120 & 100 \\
\hline
\end{tabular}
\label{tab-train-validation-nums}
\end{table}

%% file: tables/table-parameter-threshold.tex
% \begin{table}
%     \centering
%     % \caption{Pilot experiment to determine objectness threshold $\tau$.}
%     \caption{Pilot experiment to determine threshold $\tau$.}
%     \begin{tabular}{cccccc} \\
% \toprule
% $\tau$ & 0 & 0.025 & 0.05 & 0.075 & 0.1 \\
% \midrule
% Score & 530 & 884 & 923 & 879 & 866 \\
% \bottomrule
%     \end{tabular}
%     \label{tab:threshold}
% \end{table}

\begin{table}
\centering
\caption{Pilot experiment to determine threshold $\tau$.}
\begin{tabular}{cccccc} 
\hline

Detector & \multicolumn{1}{l}{$\tau$=0} & \multicolumn{1}{l}{$\tau$=0.025} & \multicolumn{1}{l}{$\tau$=0.05} & \multicolumn{1}{l}{$\tau$=0.075} & \multicolumn{1}{l}{$\tau$=0.1} \\ 
\hline
Faster R-CNN & 72 & 102 & 110 & \textbf{114} & 109 \\
YOLOv2 & 101 & 280 & \textbf{285} & 277 & 271 \\ 
YOLOv4 & 67 & 128 & \textbf{136} & 131 & 126 \\ 
YOLOR & 132 & 281 & \textbf{289} & 279 & 265 \\ 
YOLOv8 & 137 & 262 & \textbf{274} & 257 & 251 \\

% $\tau$ & 0 & 0.025 & 0.05 & 0.075 & 0.1 \\
% \midrule
% Score & 530 & 884 & 923 & 879 & 866 \\
\hline
    \end{tabular}
    \label{tab:threshold}
    % \vspace{-10pt}
\end{table}

%% file: tables/table-parameter-patchsize.tex
\begin{table*}

\centering
\caption{
\ifshepherd
\revbegin{A}{3}
\sethlcolor{yellow}\hl{
Pilot experiment to determine defensive patches size (``CA'' for canary and ``WD'' for woodpecker).
}
\revend{A}{3}
\else {Pilot experiment to determine defensive patches size (``CA'' for canary and ``WD'' for woodpecker).}
\fi
}

% \resizebox{\linewidth}{!}
{

\ifshepherd
\begin{tabular}{cc >{\columncolor{yellow}}cccccccc >{\columncolor{yellow}}c >{\columncolor{yellow}}c} 
\else 
\begin{tabular}{cccccccccccc} 
\fi

% \begin{tabular}{cc >{\columncolor{yellow}}cccccccc >{\columncolor{yellow}}c >{\columncolor{yellow}}c} 

\hline
\multirow{2}{*}{Detector} & \multirow{2}{*}{\begin{tabular}[c]{@{}c@{}}Defensive\\ Patch\end{tabular}} & \multicolumn{10}{c}{Patch size} \\ 
\cline{3-12}
 & & 20$\times$20 & 40$\times$40 & 60$\times$60 & 80$\times$80 & 100$\times$100 & 120$\times$120 & 140$\times$140 & 160$\times$160 & 180$\times$180 & 200$\times$200  \\ 
\hline
\multirow{2}{*}{\begin{tabular}[c]{@{}c@{}}Faster\\ R-CNN\end{tabular}} & CA & \clrcell{yellow}0.788 & \clrcell{yellow}0.780 & \clrcell{yellow}0.774 & 0.803 & 0.849 & \textbf{0.883} & 0.794 & \clrcell{yellow}0.258 & 0.480 & 0.303 \\
 & WD & \clrcell{yellow}0.271 & \clrcell{yellow}0.548 & \clrcell{yellow}0.568 & 0.694 & 0.749 & \textbf{0.848} & 0.772 & \clrcell{yellow}0.745 & 0.745 & 0.620 \\ 
\hline
\multirow{2}{*}{YOLOv2}  & CA & 0.897 & 0.936 & \textbf{0.962} & 0.938 & 0.931 & \clrcell{yellow}0.931 & \clrcell{yellow}0.926 & \clrcell{yellow}0.930 & 0.917 & 0.905 \\
 & WD & 0.485 & 0.704 & \textbf{0.906} & 0.866 & 0.854 & \clrcell{yellow}0.854 & \clrcell{yellow}0.852 & \clrcell{yellow}0.889 & 0.852 & 0.832 \\ 
\hline
\multirow{2}{*}{YOLOv4}  & CA & 0.885 & 0.853 & \textbf{0.888} & 0.830 & 0.805 & \clrcell{yellow}0.039 & \clrcell{yellow}0.262 & \clrcell{yellow}0.148 & 0.113 & 0.295 \\
 & WD & 0.786 & 0.803 & \textbf{0.841} & 0.805 & 0.814 & \clrcell{yellow}0.800 & \clrcell{yellow}0.750 & \clrcell{yellow}0.667 & 0.529 & 0.438 \\ 
\hline
\multirow{2}{*}{YOLOR}  & CA & 0.935 & \clrcell{yellow}0.939 & \clrcell{yellow}0.939 & \clrcell{yellow}0.943 & 0.932 & \textbf{0.949} & 0.906 & 0.845 & 0.927 & 0.924 \\
 & WD & 0.504 & \clrcell{yellow}0.616 & \clrcell{yellow}0.786 & \clrcell{yellow}0.844 & 0.917 & 0.944 & \textbf{0.954} & 0.951 & 0.921 & 0.927 \\ 
\hline
\multirow{2}{*}{YOLOv8}  & CA & 0.880 & 0.948 & 0.961 & \textbf{0.986} & 0.976 & \clrcell{yellow}0.925 & \clrcell{yellow}0.888 & \clrcell{yellow}0.852 & 0.802 & 0.715 \\
 & WD & 0.726 & 0.755 & 0.817 & \textbf{0.873} & 0.856 & \clrcell{yellow}0.865 & \clrcell{yellow}0.859 & \clrcell{yellow}0.852 & 0.852 & 0.846 \\
\hline
\end{tabular}
}

\label{tab-patch-size}
\end{table*}

%% file: tables/table-compare-all-new0918.tex
\begin{table*}
\centering

\caption{ 
% Detection performance on different object detectors targeting various adversarial patch attacks. The bold \textcolor{red}{red} F1 score denotes our method outperforms all compared methods.
\ifshepherd
\conbegin{2}
\revbegin{B}{9}
\revend{D}{4}
\sethlcolor{yellow}\hl{
Detection performance  (F1 Score) on different object detectors targeting various adversarial patch attacks. The bold red F1 score denotes our method outperforms all compared methods. FSRCNN is short of Faster R-CNN.
}
\revend{D}{4}
\revend{B}{9}
\conend{2}
\else {Detection performance (F1 Score) on different object detectors targeting various adversarial patch attacks. The bold \textcolor{red}{red} F1 score denotes our method outperforms all compared methods. FSRCNN is short of Faster R-CNN.}
\fi
}

\resizebox{\textwidth}{!}
{

\ifshepherd
\begin{tabular}{ccccccccc >{\columncolor{yellow}}cccc} 
\else 
\begin{tabular}{ccccccccccccc} 
\fi
    \hline
    
    \makecell{Attack\\ Type} & \makecell{Attack\\ Method} & Detector & Dataset  & \#Adv & LGS   & UDF   & DetectorGuard & ObjectSeeker & SAC& Mode \#1& Mode \#2& Mode \#3 \\ 
    \hline
    \multirow{8}{*}{\makecell{Known\\Attacks}} & \multirow{5}{*}{AdvPatch} & YOLOv2   & Digital  & 1242   & 0.585 & 0.519 & 0.305   & 0.720  & \textbf{0.952} & \textcolor{red}{\textbf{0.954}} & 0.882   & 0.951 \\
       &   & YOLOv2   & Physical & 280 & 0.841 & 0.809 & 0.328   & 0.929  & 0.000 & \textcolor{red}{\textbf{0.995}} & \textcolor{red}{\textbf{0.993}} & \textcolor{red}{\textbf{0.998}}  \\
       &   & YOLOv4   & Digital  & 147 & 0.790 & 0.767 & 0.251   & 0.737  & 0.487 & \textcolor{red}{\textbf{0.797}} & \textcolor{red}{\textbf{0.931}} & \textcolor{red}{\textbf{0.945}}  \\
       &   & YOLOR & Digital  & 550 & 0.803 & 0.757 & 0.137   & 0.916  & 0.697 & \textcolor{red}{\textbf{0.927}} & \textcolor{red}{\textbf{0.939}} & \textcolor{red}{\textbf{0.952}}  \\
       &   & YOLOv8   & Digital  & 819 & 0.493 & 0.783 & 0.292   & 0.841  & \textbf{0.996} & 0.974   & 0.885   & 0.971 \\ 
    \cline{2-13}
       & \multirow{2}{*}{TC-EGA} & FSRCNN   & Digital  & 87 & 0.235 & 0.475 & 0.231   & 0.702  & 0.311 & \textcolor{red}{\textbf{0.821}} & 0.609   & \textcolor{red}{\textbf{0.828}}  \\
       &   & FSRCNN   & Physical & 280 & 0.188 & 0.782 & 0.336   & 0.993  & 0.560 & \textcolor{red}{\textbf{1.000}} & 0.831   & \textcolor{red}{\textbf{1.000}}  \\ 
    \cline{2-13}
\clrrow{yellow}       & \multicolumn{3}{c}{Summary}   & 3405   & 0.610 & 0.689 & 0.278   & 0.814  & 0.828 & \textcolor{red}{\textbf{0.953}} & \textbf{\textcolor{red}{0.895}} & \textcolor{red}{\textbf{0.961}}  \\ 
    \hline
    \multirow{20}{*}{\makecell{ Unknown\\Attacks}} & \multirow{5}{*}{TC-EGA} & YOLOv2   & Digital  & 1023   & 0.582 & 0.661 & 0.329   & 0.636  & 0.023 & \textcolor{red}{\textbf{0.951}} & \textbf{\textcolor{red}{0.879}} & \textcolor{red}{\textbf{0.949}}  \\
       &   & YOLOv2   & Physical & 280 & 0.803 & 0.819 & 0.323   & 0.873  & 0.000 & \textcolor{red}{\textbf{0.972}} & \textbf{\textcolor{red}{0.987}} & \textcolor{red}{\textbf{1.000}}  \\
       &   & YOLOv4   & Digital  & 25 & 0.537 & 0.706 & 0.267   & 0.711  & 0.267 & \textcolor{red}{\textbf{0.732}} & \textbf{\textcolor{red}{0.864}} & \textcolor{red}{\textbf{0.960}}  \\
       &   & YOLOR & Digital  & 57 & 0.646 & 0.610 & 0.222   & 0.796  & 0.000 & \textcolor{red}{\textbf{0.812}} & 0.659   & \textcolor{red}{\textbf{0.899}}  \\
       &   & YOLOv8   & Digital  & 183 & 0.776 & 0.776 & 0.171   & 0.793  & 0.063 & \textcolor{red}{\textbf{0.807}} & \textbf{\textcolor{red}{0.811}} & \textcolor{red}{\textbf{0.900}}  \\ 
    \cline{2-13}
       & \multirow{7}{*}{Naturalistic}   & FSRCNN   & Digital  & 152 & 0.164 & 0.447 & 0.389   & 0.718  & 0.061 & \textcolor{red}{\textbf{0.865}} & \textbf{\textcolor{red}{0.778}} & \textcolor{red}{\textbf{0.873}}  \\
       &   & FSRCNN   & Physical & 280 & 0.545 & 0.873 & 0.409   & 0.958  & 0.000 & \textcolor{red}{\textbf{0.995}} & 0.796   & \textcolor{red}{\textbf{0.995}}  \\
       &   & YOLOv2   & Digital  & 587 & 0.665 & 0.850 & 0.519   & 0.653  & 0.030 & \textcolor{red}{\textbf{0.932}} & \textbf{\textcolor{red}{0.878}} & \textcolor{red}{\textbf{0.934}}  \\
       &   & YOLOv2   & Physical & 280 & 0.181 & 0.843 & 0.431   & 0.903  & 0.000 & \textcolor{red}{\textbf{0.993}} & \textbf{\textcolor{red}{0.927}} & \textcolor{red}{\textbf{0.996}}  \\
       &   & YOLOv4   & Digital  & 114 & 0.682 & 0.670 & 0.201   & 0.711  & 0.314 & \textcolor{red}{\textbf{0.764}} & \textbf{\textcolor{red}{0.840}} & \textcolor{red}{\textbf{0.894}}  \\
       &   & YOLOR & Digital  & 75 & 0.496 & 0.619 & 0.294   & 0.746  & 0.000 & \textcolor{red}{\textbf{0.803}} & 0.707   & \textcolor{red}{\textbf{0.887}}  \\
       &   & YOLOv8   & Digital  & 424 & 0.532 & 0.661 & 0.846   & 0.566  & 0.055 & \textcolor{red}{\textbf{0.871}} & \textbf{\textcolor{red}{0.850}} & \textcolor{red}{\textbf{0.911}}  \\ 
    \cline{2-13}
       & \multirow{7}{*}{UPC}& FSRCNN   & Digital  & 65 & 0.263 & 0.432 & 0.800   & 0.719  & 0.086 & \textcolor{red}{\textbf{0.855}} & 0.725   & \textcolor{red}{\textbf{0.864}}  \\
       &   & FSRCNN   & Physical & 280 & 0.586 & 0.750 & 0.341   & 0.937  & 0.000 & \textcolor{red}{\textbf{0.998}} & 0.850   & \textcolor{red}{\textbf{0.998}}  \\
       &   & YOLOv2   & Digital  & 807 & 0.679 & 0.827 & 0.724   & 0.603  & 0.007 & \textcolor{red}{\textbf{0.915}} & \textbf{\textcolor{red}{0.889}} & \textcolor{red}{\textbf{0.938}}  \\
       &   & YOLOv2   & Physical & 280 & 0.761 & 0.965 & 0.589   & 0.800  & 0.000 & \textcolor{red}{\textbf{0.989}} & 0.941   & \textcolor{red}{\textbf{0.998}}  \\
       &   & YOLOv4   & Digital  & 91 & 0.711 & 0.756 & 0.675   & 0.734  & 0.333 & \textcolor{red}{\textbf{0.779}} & \textbf{\textcolor{red}{0.910}} & \textcolor{red}{\textbf{0.950}}  \\
       &   & YOLOR & Digital  & 201 & 0.661 & 0.809 & 0.745   & 0.866  & 0.000 & 0.855   & 0.846   & \textcolor{red}{\textbf{0.934}}  \\
       &   & YOLOv8   & Digital  & 151 & 0.731 & 0.781 & 0.644   & 0.664  & 0.064 & \textcolor{red}{\textbf{0.936}} & \textbf{\textcolor{red}{0.886}} & \textcolor{red}{\textbf{0.950}}  \\ 
    \cline{2-13}
    \clrrow{yellow} & \multicolumn{3}{c}{Summary}   & 5355   & 0.617 & 0.769 & 0.527   & 0.716  & 0.037 & \textcolor{red}{\textbf{0.926}} & \textbf{\textcolor{red}{0.873}} & \textcolor{red}{\textbf{0.947}}  \\ 
    \hline
     \multicolumn{4}{c}{\bf Total} & 8760   & 0.622 & 0.749 & 0.458   & 0.807  & 0.445 & \textcolor{red}{\textbf{0.948}} & \textbf{\textcolor{red}{0.885}} & \textcolor{red}{\textbf{0.964}}  \\
    \hline
    \end{tabular}

}

\label{tab-compare-all}
\end{table*}

%% file: tables/table-false-positive.tex
\begin{table*}
\centering

\caption{ 
% Detection performance on different object detectors targeting various adversarial patch attacks. The bold \textcolor{red}{red} F1 score denotes our method outperforms all compared methods.
\ifshepherd
\conbegin{2}\revbegin{B}{9}\revbegin{D}{2}
\sethlcolor{yellow}\hl{
False positive rates targeting various adversarial patch attacks.
}
\revend{D}{2}\revend{B}{9}\conend{2}
\else {False positive rates targeting various adversarial patch attacks.}
\fi
}

\resizebox{\textwidth}{!}
{

    \ifshepherd
        \begin{tabular}{c >{\columncolor{yellow}}c >{\columncolor{yellow}}c >{\columncolor{yellow}}c >{\columncolor{yellow}}c >{\columncolor{yellow}}c >{\columncolor{yellow}}c >{\columncolor{yellow}}c >{\columncolor{yellow}}c >{\columncolor{yellow}}c >{\columncolor{yellow}}c} 
    \else 
        \begin{tabular}{ccccccccccc} 
    \fi
    \hline
    
    \begin{tabular}[c]{@{}c@{}}Attack Type\end{tabular} & \begin{tabular}[c]{@{}c@{}}Attack Method\end{tabular} & \#Distinct Benign & LGS & UDF & DetectorGuard & ObjectSeeker & SAC & Mode \#1 & Mode \#2 & Mode \#3\\ 
    \hline
    
    \multirow{3}{*}{\begin{tabular}[c]{@{}c@{}}Known\\ Attacks\end{tabular}} & AdvPatch & 3038 & 0.066 & 0.065 & 0.161& 0.320 & 0.002 & 0.057& 0.017& 0.066 \\
    
    & TC-EGA & 367& 0.008 & 0.022 & 0.292& 0.134 & 0.000 & 0.033& 0.005& 0.033 \\ 
    \cline{2-11}
    
    & Summary& 3405 & 0.060 & 0.060 & 0.175& 0.300 & 0.001 & 0.054& 0.016& 0.062 \\ 
    \hline
    
    \multirow{4}{*}{\begin{tabular}[c]{@{}c@{}}Unknown\\ Attacks\end{tabular}} & TC-EGA & 1519 & 0.047 & 0.047 & 0.333& 0.409 & 0.002 & 0.066& 0.018& 0.076 \\
    
    & Naturalistic& 1305 & 0.084 & 0.070 & 0.314& 0.359 & 0.005 & 0.067& 0.024& 0.078 \\
    
    & UPC & 1372 & 0.058 & 0.060 & 0.324& 0.388 & 0.007 & 0.069& 0.017& 0.077 \\ 
    \cline{2-11}
    
    & Summary& 2855 & 0.067 & 0.060 & 0.373& 0.368 & 0.005 & 0.063& 0.019& 0.072 \\ 
    \hline
    
    \multicolumn{2}{c}{\bf Total}& 6071 & 0.058 & 0.053 & 0.155& 0.282 & 0.004 & 0.049& 0.014& 0.055 \\
    \hline
    
    \end{tabular}
}

\label{tab-false-positive}
\end{table*}

%% file: tables/table-target-car.tex
\begin{table*}
% \caption{The performance in detecting object-specific attacks.}
\caption{
\conbegin{1/2}
\revbegin{B}{9}
\ifshepherd
\hl{The performance in detecting object-specific attacks.}
\else {The performance in detecting object-specific attacks.}
\fi
\revend{B}{9}
\conend{1/2}
}
\centering
% \resizebox{\textwidth}{!}
{

    \ifshepherd
        \begin{tabular}{ccccccc >{\columncolor{yellow}}cccc} 
    \else 
        \begin{tabular}{ccccccccccc} 
    \fi
    \hline
    
    Target & Dataset & \#Adv & LGS & UDF & DetectorGuard & ObjectSeeker & SAC & Mode \#1 & Mode \#2 & Mode \#3 \\ 
    \hline
     
     Car & BDD100K & 100 & 0.529 & 0.652 & 0.164 & 0.802 & 0.496 & 0.954 & 0.920 & 0.965 \\
     
     \clrrow{yellow} Bicycle & VOC07 & 64 & 0.361 & 0.590 & 0.517 & 0.673 & 0.476 & 0.920 & 0.852 & 0.920 \\
     
     \clrrow{yellow} Dog & VOC07 & 96 & 0.107 & 0.256 & 0.317 & 0.540 & 0.769 & 0.839 & 0.606 & 0.830 \\
     
     \clrrow{yellow} Train & VOC07 & 78 & 0.245 & 0.255 & 0.261 & 0.685 & 0.836 & 0.846 & 0.683 & 0.870 \\
    \hline
     
    \clrrow{yellow} \multicolumn{2}{c}{\bf Total} & 338 & 0.337 & 0.467 & 0.307 & 0.693 & 0.667 & \textbf{\textcolor{red}{0.888}} & \textbf{\textcolor{red}{0.772}} & \textbf{\textcolor{red}{0.893}} \\

     \hline
 \end{tabular}
 
}
\label{tab-target-car}
\end{table*}

%% file: tables/table-sample-specific.tex
\begin{table*}
\centering
% \caption{The performance in detecting sample-specific attacks.}
\caption{
\conbegin{2}
\revbegin{B}{9}
\ifshepherd
\sethlcolor{yellow}\hl{The performance in detecting sample-specific attacks.}
\else 
{The performance in detecting sample-specific attacks.}
\fi
\revend{B}{9}
\conend{2}
}
% \resizebox{\textwidth}{!}
{

    \ifshepherd
        \begin{tabular}{cccccc >{\columncolor{yellow}}cccc} 
    \else 
        \begin{tabular}{cccccccccc} 
    \fi
    \hline
    
    Dataset & \#Adv & LGS & UDF & DetectorGuard & ObjectSeeker & SAC & Mode \#1 & Mode \#2 & Mode \#3\\ 
    \hline
    
    VOC07 & 25& 0.880 & 1.000 & 0.400& 0.898& 0.438 & 1.000 & 1.000 & 1.000  \\
    
    COCO& 25& 0.941 & 0.962 & 0.323& 0.857& 0.077 & 1.000 & 1.000 & 1.000  \\
    
    Inria & 25& 0.936 & 0.960 & 0.375& 0.939& 0.387 & 1.000 & 0.980 & 1.000  \\
    
    Physical& 25& 0.936 & 0.980 & 0.077& 0.885& 0.438 & 1.000 & 1.000 & 1.000  \\
    \hline

    \clrrow{yellow} \textbf{Total} & 100 & 0.923 & 0.975 & 0.306 & 0.894 & 0.347 & \textcolor{red}{\textbf{1.000}} & \textcolor{red}{\textbf{0.995}} & \textcolor{red}{\textbf{1.000}}  \\ 

    \hline
    
    \end{tabular}

}
\label{tab-sample-specific}
\end{table*}

%% file: tables/table-efficiency.tex
\begin{table*}[t]
\centering
\caption{
\ifshepherd
{
\conbegin{2}\revbegin{B}{5/9}
\sethlcolor{yellow}\hl{
Efficiency (time unit: milliseconds/image; ``A'' for adversarial samples and ``B'' for benign samples).
}\revend{B}{5/9}\conend{2}
}
\else {Efficiency (time unit: milliseconds/image; ``A'' for adversarial samples and ``B'' for benign samples).}
\fi
}

\resizebox{\linewidth}{!}{

    \ifshepherd
    \begin{tabular}{ccrrrrrrrr>{\columncolor{yellow}}r>{\columncolor{yellow}}rrrrrrr} 
    \else 
    \begin{tabular}{ccrrrrrrrrrrrrrrrr} 
    \fi
    \hline

    \multirow{2}{*}{Detector} & \multirow{2}{*}{Dataset} & \multicolumn{2}{c}{LGS} & \multicolumn{2}{c}{UDF} & \multicolumn{2}{c}{DetectorGuard} & \multicolumn{2}{c}{ObjectSeeker} & \multicolumn{2}{c}{\clrcell{yellow} SAC} & \multicolumn{2}{c}{Mode
     \#1} & \multicolumn{2}{c}{Mode
     \#2} & \multicolumn{2}{c}{Mode
     \#3} \\ 
    \cline{3-18}
    
     & & A & B & A & B & A & B & A & B & A & B & A & B & A & B & A & B \\ 
    \hline
    
     & VOC07 & 34.38 & 34.16 & 42.33 & 42.05 & 244.67 & 243.31 & 3711.02 & 3704.55 & 41.47 & 36.95 & 62.44 & 45.49 & 52.92 & 45.49 & 98.81 & 53.63 \\
    \clrrow{yellow}\clrcell{white}     & COCO & 29.79 & 39.74 & 37.54 & 35.25 & 216.32 & 214.32 & 4144.83 & 4223.96 & 49.67 & 52.77 & 61.70 & 52.54 & 56.21 & 52.58 & 82.07 & 77.10 \\
    \clrrow{yellow}\clrcell{white} & Inria & 50.76 & 45.81 & 44.59 & 35.32 & 212.29 & 264.60 & 3789.28 & 3774.82 & 58.98 & 51.85 & 62.99 & 54.14 & 64.25 & 52.57 & 94.93 & 93.16 \\
    \clrrow{yellow}\clrcell{white} & Indoor & 46.17 & 45.78 & 38.36 & 38.06 & 227.43 & 247.56 & 4262.78 & 5037.56 & 55.16 & 55.34 & 73.55 & 41.18 & 63.46 & 42.47 & 75.88 & 59.40 \\
    \clrrow{yellow}\clrcell{white} \multirow{-5}{*}{\makecell{Faster\\ R-CNN}} & Outdoor & 36.18 & 45.78 & 38.92 & 38.61 & 261.55 & 248.17 & 4306.34 & 4282.34 & 46.69 & 44.79 & 79.11 & 46.47 & 67.56 & 46.24 & 84.23 & 50.42 \\ 
    \hline
    
    & VOC07 & 44.60 & 44.46 & 48.48 & 40.16 & 373.79 & 372.55 & 3825.37 & 3845.85 & 61.97 & 50.77 & 103.73 & 61.05 & 97.47 & 58.04 & 116.62 & 92.42 \\
    \clrrow{yellow}\clrcell{white} & COCO & 39.60 & 40.19 & 41.33 & 43.67 & 375.21 & 372.91 & 3731.84 & 3569.69 & 68.47 & 72.67 & 123.76 & 75.82 & 101.21 & 75.47 & 135.79 & 86.28 \\
    \clrrow{yellow}\clrcell{white} & Inria & 44.95 & 45.45 & 39.66 & 39.49 & 386.01 & 389.60 & 3736.74 & 3704.29 & 116.78 & 110.07 & 107.07 & 70.03 & 107.07 & 49.88 & 120.99 & 84.96 \\
    \clrrow{yellow}\clrcell{white} & Indoor & 53.75 & 52.41 & 44.28 & 44.34 & 380.37 & 374.70 & 3975.72 & 3910.25 & 79.91 & 79.29 & 70.03 & 47.42 & 92.42 & 46.99 & 93.63 & 84.25 \\
    \clrrow{yellow}\clrcell{white} \multirow{-5}{*}{YOLOv2} & Outdoor & 54.11 & 54.84 & 45.91 & 46.30 & 366.70 & 377.12 & 3996.66 & 3942.15 & 83.20 & 77.94 & 95.60 & 45.91 & 96.43 & 44.31 & 134.41 & 81.70 \\ 
    \hline
    
     & VOC07 & 57.88 & 48.63 & 47.43 & 43.00 & 177.01 & 176.11 & 4274.52 & 4285.10 & 52.94 & 41.07 & 110.21 & 86.73 & 114.59 & 85.91 & 143.49 & 81.94 \\
    \clrrow{yellow}\clrcell{white} & COCO & 43.03 & 40.06 & 43.97 & 49.57 & 179.57 & 178.19 & 4214.07 & 4244.60 & 41.26 & 46.36 & 113.51 & 62.93 & 112.10 & 62.66 & 133.61 & 76.01 \\
    \clrrow{yellow}\clrcell{white} \multirow{-3}{*}{YOLOv4}& Inria & 47.28 & 49.25 & 40.48 & 42.76 & 194.06 & 183.18 & 4121.41 & 4191.12 & 59.67 & 63.14 & 114.05 & 68.97 & 112.98 & 69.16 & 138.89 & 88.53 \\ 
    \hline
    
     & VOC07 & 78.29 & 77.39 & 95.74 & 86.79 & 1262.27 & 1259.97 & 3999.51 & 4062.69 & 142.37 & 151.65 & 119.24 & 74.18 & 116.01 & 72.89 & 126.06 & 106.50 \\
    \clrrow{yellow}\clrcell{white} & COCO & 86.00 & 87.76 & 88.66 & 90.55 & 1260.52 & 1274.97 & 4103.94 & 4116.97 & 159.35 & 161.02 & 124.47 & 79.05 & 120.26 & 78.06 & 128.57 & 109.77 \\
    \clrrow{yellow}\clrcell{white} \multirow{-3}{*}{YOLOR}& Inria & 91.60 & 93.10 & 88.34 & 94.13 & 1270.56 & 1254.00 & 4098.86 & 4142.22 & 303.32 & 288.21 & 125.29 & 100.60 & 124.74 & 98.62 & 123.91 & 114.03 \\ 
    \hline
    
     & VOC07 & 72.87 & 50.29 & 61.76 & 59.92 & 190.37 & 187.89 & 1374.46 & 1378.66 & 77.74 & 67.45 & 77.90 & 74.01 & 75.80 & 73.12 & 79.95 & 75.13 \\
    \clrrow{yellow}\clrcell{white} & COCO & 73.69 & 54.14 & 42.07 & 62.41 & 189.71 & 188.31 & 1400.48 & 1409.79 & 71.72 & 74.10 & 72.38 & 69.39 & 72.55 & 69.47 & 76.47 & 71.23 \\
    \clrrow{yellow}\clrcell{white} \multirow{-3}{*}{YOLOv8}& Inria & 86.55 & 79.56 & 70.11 & 60.14 & 191.41 & 194.92 & 1420.11 & 1433.25 & 109.82 & 102.25 & 72.85 & 66.46 & 73.02 & 66.44 & 77.24 & 68.45 \\
    \hline

    \end{tabular}

}
\label{tab-runtime}
\end{table*}

%% file: tables/table-adaptive-attack.tex
\begin{table*}[t]
\centering
\caption{
\ifshepherd
\revbegin{A}{4}\hl{Defense performance and time cost for adaptive attacks.}\revend{A}{4}
\else {Defense performance and time cost for adaptive attacks.}
\fi
}
\ifshepherd
\begin{tabular}{c >{\columncolor{yellow}}c >{\columncolor{yellow}}c >{\columncolor{yellow}}c >{\columncolor{yellow}}c >{\columncolor{yellow}}c >{\columncolor{yellow}}cc >{\columncolor{yellow}}c >{\columncolor{yellow}}c >{\columncolor{yellow}}c >{\columncolor{yellow}}c >{\columncolor{yellow}}c}
\else 
\begin{tabular}{ccccccccccccc}
\fi
\hline
\multirow{2}{*}{Dataset} & & \multicolumn{5}{c}{Number of samples  successfully bypassing defense methods} & \multirow{2}{*}{} &\multicolumn{5}{c}{Time cost for generating adaptive attack patches (in hours)} \\ 
\cline{3-7}\cline{9-13}

& \multirow{-2}{*}{\#Adv} & Dep \#1 & Dep \#2 & Dep \#3 & Dep \#4 & UDF & & Dep \#1 & Dep \#2 & Dep \#3 & Dep \#4 & UDF \\ \hline

VOC07 & 105 & 71 & \textbf{0} & 62 & \textbf{0} & 90 & & 79.75 & 200.27 & 70.25 & 86.91 & 23.92 \\

COCO & 128 & 87 & \textbf{0} & 71 & \textbf{0} & 87 & & 100.67 & 234.43 & 88.10 & 118.67 & 21.77 \\

Inria & 135 & 93 & \textbf{0} & 71 & \textbf{0} & 87 & & 102.94 & 241.13 & 93.98 & 124.79 & 30.16 \\

Physical & 120 & 87 & \textbf{0} & 24 & \textbf{0} & 93 & & 97.76 & 230.61 & 80.64 & 103.19 & 28.80 \\ \hline

Total & 488 & 338 & \textbf{0} & 228 & \textbf{0} & 357 & & 381.12 & 906.44 & 332.97 & 433.56 & 104.65 \\ 
\hline
\end{tabular}

\label{tab-adaptive-attack}
\end{table*}

%% file: Discussion.tex
\section{Discussion and Limitations}\label{section-discussion}

% \textbf{Potential Bypass Techniques.} In object detection tasks, objectness scores are used to determine the possibility of an object being presented in a given image. We determine candidate positions to place the defensive patches based on potential objects of interest class with low objectness scores. However, the attacker may further make these candidate locations indeterminate. To counter such attacks, we think that there are two potential measures. The first measure involves randomization, which can either be a random selection of defensive patches or random positions of defensive patches. The second measure involves adjusting the threshold of objectness even further. This makes it more difficult for attackers to decrease the objectness scores of the victim objects. While it is true that no defense method is foolproof, the proposed method has proven to be effective and robust in practice. It raises the bar for attackers by making it more difficult for them to launch successful attacks.

\textbf{Overlap with Adversarial Patches.} 
The overlap of woodpecker and attack patches is not a key factor for defending adversarial patch attacks.
% Our woodpecker patches detect the attacks by interacting with the adversarial patches. 
We conduct experiments with other types of content, e.g., blank, black, gray and noise blocks, to demonstrate that simply overlapping with attack patches is ineffective in recovering the victim object. The result is shown in Figure \ref{fig-compare-wd}. 
Only the woodpecker counteracts the attack patch and recovers the person. All other block patches do not impact the effect of adversarial patches. 
Furthermore, woodpecker can achieve the same defense effectiveness by covering only a tiny part or even without any overlap of the attack patch, as shown in Figure~\ref{fig-occlusion-wd}(a) and (b), respectively. 
% \lb{
In summary, the specially crafted content of woodpecker makes it effective in detecting adversarial patch attacks regardless of whether it overlaps with the adversarial patches or not.
% }
% In summary, the proposed woodpecker works based on its interaction with attack patches. 
% This makes it effective in detecting adversarial patch attacks regardless of whether it overlaps with the adversarial patches or not.

% Similarly, as shown in Figure \ref{fig-occlusion-cawd} in Appendix \ref{appendx:overlapping}, the woodpecker patch covers only a small part of the attack patch, and it can weaken the interference of the attack patches. Then experiments were conducted with other types of content, e.g., blank, black, gray, and noise blocks, to further demonstrate the effectiveness of the proposed defensive patches, as shown in Figure \ref{fig-compare-ca} and Figure \ref{fig-compare-wd}. Even if the canary patch does not totally cover the adversarial patches, the canary patch is still affected and not detected. Similarly, after importing woodpecker, the victim object is newly detected. In contrast, other patches cannot be used to detect adversarial attacks. In summary, the proposed defensive patches work based on their interaction with attack patches. This makes them effective in detecting adversarial patch attacks regardless of whether they overlap with the adversarial patches or not.

\revbegin{B}{6}
\hjj{
\textbf{False Positives and False Negatives.} 
Canary, as a fragile object, may not be accurately detected in scenarios with complex background, resulting in false positives. In some other scenarios, the perturbation effect of the adversarial patch may be not enough to affect the canary, causing false negatives (FNs). 
For woodpecker, if the attacked target is inherently difficult to recognize and has been heavily compromised by the adversarial patch, the woodpecker may fail to recover it (FN); and if a nearby area is identified to possibly include a target object by the detector, woodpecker may augment the information and lead to a FP.
Fortunately, in many cases, canaries and woodpeckers can complement each other effectively. Additionally, we believe that using a more comprehensive training set to generate canaries and woodpeckers will further reduce FPs and FNs. 
Appendix~\ref{appendix:FN FP} presents a few examples of FPs and FNs.
}
\revend{B}{6}

\revbegin{B}{8}
\hjj{
\textbf{Effect of Training Dataset.}
To some extent, our method is independent of the training dataset. 
However, in theory, it can achieve better performance to train the patches using the samples closely related to the usage scenario, which may further improve the performance compared with the usage in the paper. Such a training is feasible in some cases, e.g., defending against attacks for a specific surveillance camera. We have demonstrated it in our physical-world dataset. For each indoor or outdoor scenario, we randomly choose 20 samples and generate corresponding adversarial samples for the four attacks targeting YOLOv2. All the three detection modes (\textit{canary only}, \textit{woodpecker only}, \textit{canary + woodpecker}) achieve 100\% precision for all cases and the third mode even achieves an F1 of 100\%.
}
\revend{B}{8}

\revbegin{C}{1}
\lb{
\textbf{Threshold $\tau$.}
To enable the detection of unknown attacks, we aim for the threshold $\tau$ to be independent of the attack methods. We generated 100 validation samples for each of the four attack methods (AdvPatch, TC-EGA, Naturalistic, and UPC) and conducted experiments using YOLOv2 to validate this. The results indicate that for all four attack methods, the suitable $\tau$ is consistently 0.05. This suggests that for YOLOv2, the threshold remains independent of the attack methods.
}
\revend{C}{1}

\revbegin{C}{2}
\hjj{
\textbf{Candidate Box around Non-Attacked Objects.} 
It is theoretically unavoidable to find candidate box around a non-attacked object, but such scenarios do not cause serious FPs. For inconspicuous objects in the input that are not detected by target detection models, such as a small, blurry part of a person, the objectness score is typically very low, which can result in the detection of a candidate box. 
To this end, we conducted a dedicated statistical analysis. It is shown that out of 6,071 experimental samples, 15.5\% contained candidate boxes for non-attacked objects, only resulting in a small number of FPs (4.9\% and 1.4\% for canary and woodpecker, respectively).
}
\revend{C}{2}

\revbegin{B}{1}
\lb{\textbf{Adaptive Parameters.}
As a learning-based approach, our method is sensitive to parameter tuning. Ideally, although challenging, it would be advantageous for the key parameters to be adaptive. This represents a valuable avenue for future research. We conducted preliminary experiments that adjusted the size of the woodpecker based on the size of the candidate boxes. Experiments targeting the AdvPatch attack showed that applying a larger woodpecker to larger candidate boxes can further enhance performance, resulting in an approximate 2\% increase in the F1 score. The intuition behind this is that larger concealed objects require a correspondingly larger woodpecker for effective recovery.}
\revend{B}{1}

\revbegin{D}{3}
\hjj{
\textbf{Multiple Adversarial Patches.}
We have studied how resilient of our approach to multiple adversarial patches. 150 adversarial samples have been tested, with 3 to 8 adversarial patches. The result shows that, our approach can effectively detect adversarial attacks in 149 the samples, outperforming the five comparative methods.
}
\revend{D}{3}

\revbegin{A}{1}
\lb{
\textbf{Transferability.}
We conduct experiments on the transferability of defense patches across three object types: \textit{person}, \textit{car} and \textit{bicycle}. The results indicate that the defensive patches exhibit good transferability in some scenarios. For example, the car-specific defense patches can detect adversarial samples that conceal persons, achieving performance almost equivalent to that of the person-specific canary and woodpecker. However, in some cases, performance significantly declined. For example, when applying a person-specific canary to bicycle samples, the F1 score decreased from 0.920 to 0.786. 
Furthermore, we perform joint optimization using samples from the three object types, resulting in defensive patches that led to F1 score decreases of 0.029 (Mode \#1), 0.006 (Mode \#2), and 0.043 (Mode \#3) compared to the object-specific defense patches. These experiments suggest that object-specific defense patches possess a certain degree of transferability, but it is insufficient. Joint optimization shows promise for providing defensive patches with improved transferability. In practice, we recommend using object-specific defense patches when the types of protected object are fixed.
}
\revend{A}{1}

\textbf{Higher Randomness.} 
% \hjjrm{In our experimental setup, we only introduced limited randomness. For instance, during the generation of canaries, we optimized them for five initial objects at five fixed positions.} 
Although our experiments have demonstrated that limited randomness is effective enough in countering adaptive attacks,
% \revbegin{B}{4}
% \lb{
e.g., three different canaries for two positions (see \S\ref{subsec:eval:adaptive}).
% }
% \revend{B}{4}
Future endeavors would involve enhancing the level of randomness if feasible in terms of resources and time. This would encompass introducing additional random positions and initial objects, as well as conducting experiments on a larger scale. Additionally, we intend to explore other factors of randomness, such as the shape of defensive patches. We firmly believe that incorporating more randomness will further enhance the robustness of our defense mechanism.

\textbf{Efficiency.} 
% Despite the effectiveness of the proposed method, there are concerns about the time cost of the proposed approach. To address this issue, 
To reduce the time cost, 
one feasible solution 
% \hjjrm{is to utilize parallel computing techniques, which can significantly decrease the time required for the detection. Another feasible solution} 
is to use random sampling to reduce the number of images that need to be processed. In monitoring surveillance, images are continuously collected, but not every frame requires to be detected. We can improve the system's overall efficiency by randomly selecting a subset of frames for detection. %In summary, these solutions can mitigate these concerns by leveraging parallel computing and random sampling. We will investigate the potential techniques in the future.

\textbf{Adversarial Attacks for 3D Object Detection.} Adversarial attacks for 3D object detection involve placing a physically realizable adversarial object that can make some objects undetected by a 3D object detector. This type of attack shares similar principles with adversarial patch attacks for 2D object detection. Theoretically, our defensive patches can also be adapted to detect adversarial attacks for 3D object detection. We can construct a brittle 3D object and import it into the input data. Then we check the state of the imported brittle 3D objects to determine whether there has been an adversarial attack on the 3D detector.
We leave the adaptation of our approach for 3D object detection as a future research direction.

%% file: Related.tex
\section{Related Work}
An increasing number of researchers have explored security attacks
~\cite{cite_cls_pintor2023imagenet, cite_demontis2019adversarial, cite_attack_cls_2022tnt, cite_autodrive_2022using, USENIX23_person_like_car, Reza2019privacy, Tianyuan2020hybrid, Aanjhan2023mmspoof, cite_autodrive_2023vulnerability, cite_attack_cls_2023lsd, Murat2023adv, cite_autodrive_2023tpatch, cite_Sayles_2021_CVPR, cite_Ryan_space_ad, cite_Ruoyu_use, jan_aaai_19_ada}
in machine learning models and proposed various defense methods~\cite{cite_def_cls_2018defending, cite_def_cls_2023resisting, cite_def_cls_2023new, cite_def_cls_2022patchcleanser,cite_def_cls_2023nnsplitter, cite_pmlr_v162_tramer22a, cite_shan_22_post, cite_shangotta_20}.
% \lb{
The following studies are closely related to the subject of this paper,
% }

% An increasing number of researchers have explored security attacks
% in machine learning models and proposed various defense methods.

\textbf{Attacking Image Classification.} Adversarial attacks can easily fool DNN-based models. Most adversarial attacks introduce a global perturbation into the input image, which may mislead a wrong prediction. Brown et al. \cite{Brown2017AdversarialP} proposed the first adversarial patch attack method to attack image classifier models, which makes adversarial patch attacks used in the physical world by applying the patch to the victim object. Road sign recognition plays an important role in self-driving. Adversarial attacks on road signs may cause property damage or user casualties. Some researches \cite{cite_attack_cls_2022tnt, LiICMLIRoadAttack, LiuAAAIRoadAttack, liu2020clsattack, cite_attack_cls_2023lsd} studied a range of adversarial patch attack methods with various threat models.

\textbf{Attacking 2D Object Detection.} Since the predicted content of the object detection task is more complex, patch attacks against object detectors are more challenging than image classifiers. The attacker can distort the pixels in a bounded region and use different objective optimization to accomplish attack effects such as hiding attacks \cite{attack_advtext, attack_fool}and misclassification attacks \cite{attack_naturalistic}\cite{attack_upc}. Thys et al. \cite{attack_fool} proposed a printable adversarial patch to make a person can evade detection. Huang et al. \cite{attack_upc} present a Universal Physical Camouflage Attack to make objects misclassified. They masquerade the victim object with texture patterns for attacking object detectors. Hu et al. \cite{attack_naturalistic} used the manifold of generative adversarial networks (GANs) to make the adversarial patches more realistic and natural-looking. Hu et al. \cite{attack_advtext} proposed TC-EGA significantly lowered the detection performance of object detectors.

\textbf{Attacking 3D Object Detection.} DNNs have made a massive progress in 3D object detection, which is vital, especially in autonomous driving scenarios. Several studies \cite{cao2019adversarial3D, Tu3DPhyAdvEx, Xiang3DGenerateAdpoint}have demonstrated that LiDAR detectors can be attacked by introducing 3D adversarial objects or modifying the point cloud sensory data, which leads to a threat to self-driving. Xiang et al. \cite{Xiang3DGenerateAdpoint} proved that models taking point cloud data as input can be attacked. Tu et al. \cite{Tu3DPhyAdvEx} place an adversarial object on the roof of victim vehicles to escape from the LiDAR detectors. Cao et al. \cite{cao2019adversarial3D} generated robust physical adversarial objects to successfully attack the Baidu Apollo system.
% \lb{
They also presented an effective adversarial sensor perturbation attack~\cite{cite_3d_cao19xiao}
and leveraged laser-based spoofing techniques to physically remove selected 3D point clouds to hide an object of interest 
% [You Can’t See Me].
~\cite{cite_3d_cao2023you}.
Liu et al.~\cite{cite_3d_slowlidar}
presented an attack method SlowLiDAR to maximize LiDAR detection runtime.
% }

\textbf{Defending Against Patch Attacks for Image Classification.}
Most existing defenses \cite{hayes2018visible, levine2020randomized, LGS, xiang2021patchguard, Chiang2020Certified, cite_def_cls_2022patchcleanser} are designed for image classification tasks. Among them, PatchCleanser~\cite{cite_def_cls_2022patchcleanser} defends attacks by introducing supplementary information into the input. Its fundamental principle involves applying a mask (a grey block) at different positions in the input to obtain multiple (at least nine) masked samples. Subsequently, a differential analysis of their classification results is conducted to detect potential adversarial attacks, which can lead to inconsistent outputs. PatchCleanser can effectively neutralize the effect of adversarial patches and detect attacks against image classifiers. However, it introduces at least nine times the time overhead. In contrast to simple pixel masks, canary and woodpecker consist of pixels with specific purposes, thereby eliminating the need for excessive differential comparisons and resulting in limited time overhead. Additionally, canary provides passive means of probing adversarial patches, offering a more comprehensive detection and effectively raising the bar for attacks. 
% It is worth noting that PatchCleanser boasts a remarkable feature: it can restore correct classifications for certain adversarial samples through the second-round masking. Woodpecker can also recover hidden target objects but without incurring any additional time overhead.

% \textbf{Defending Against Patch Attacks for Object Detection.}
% Few defense methods~\cite{DefenseMask, DefendEnergy, LiuSegDefend, USENIX23_person_like_car, ObjectSeeker}  have been studied in the object detection models. Kim et al. \cite{DefendEnergy} analyze patch-feature energy and smooth the features of potential areas of the input image, which influence the detector's speed. Chiang et al. \cite{DefenseMask} first feed the input image into an extra MaskNet to block potential adversarial patch areas, then send the modified image into the object detector. Using an image segmentation network to shield the perturbation of adversarial patches reduces the efficiency of detecting objects. And obtaining such a segmentation network requires lots of paired benign and adversarial samples. Liu et al.\cite{LiuSegDefend} also use an extra segmentation network to ensure the safety of object detectors. Xiang et al. \cite{ObjectSeeker} eliminate the interference of the adversarial patches by continuously cropping the input image, which undisturbed the detected part of the image. However, cropping input image aimlessly and integrating scattered objection results increase object detection time.

\revbegin{D}{4}
\lb{
\textbf{Defending Against Patch Attacks for Object Detection.}
Some defense methods aim to mitigate adversarial patch attacks by minimizing the perturbations associated with potential adversarial patches. For instance, SAC \cite{LiuSegDefend} trains an additional segmentation network to identify and mask potential adversarial patch areas in the input samples, thereby enhancing its defense against attacks. Similarly, APM \cite{DefenseMask} also employs an extra network to recognize and mask adversarial patches present in the input. In contrast, APE \cite{DefendEnergy} masks features of potential areas in the input image based on the feature energy distribution within the convolutional layers. These mask-based approaches heavily rely on the characteristics of known samples and do not adequately address the potential for unseen attack methods. As demonstrated in \S\ref{EffectivenessAnalysis}, SAC experiences a significant drop in performance when confronted with unknown attack techniques.
LGS \cite{LGS} is also used to protect object detectors by treating adversarial patches as noise and mitigating their impact through input regularization gradients. However, based on our observations, certain adversarial patches, such as Naturalistic patches, cannot simply be regarded as noise, which leads to poor detection performance (see \S\ref{EffectivenessAnalysis}).
ObjectSeeker~\cite{ObjectSeeker} achieves correct identification of hidden target objects by cropping the input into multiple segments, ensuring that some segments either do not contain adversarial patches or only partially include them. ObjectSeeker does not require training and can be applied independently; however, aimlessly cropping input image and integrating scattered objection results significantly increase object detection time (see \S\ref{subsec:Efficiency}).
}

\lb{
Unlike the direct removal of perturbations caused by adversarial patches, DetectorGuard \cite{DetectorGuard} utilizes differential analysis of outputs from multiple models to detect attacks. It introduces an additional model named Objectness Predictor, which generates an objectness map that highlights potential targets in the input image. If this map fails to ``explain'' the output of the protected detector, i.e., if there are no detected objects corresponding to the objectness map, an attack is reported. The performance of DetectorGuard relies on Objectness Predictor, which can potentially be bypassed through targeted attacks. To explore the weakness, we conducted experiments to jointly optimize and train adversarial patches capable of simultaneously targeting both the detector and Objectness Predictor. The results indicate that generating adversarial patches that compromise the effectiveness of DetectorGuard is not a difficult task.
}

\lb{
UDF \cite{UDF} trains a defensive framework that is incorporated into the input to actively interfere with adversarial patches, enabling the detector to produce correct outputs. Our approach also employs an active defense strategy. While woodpecker shares a similar philosophy with UDF, canary adopts a completely different strategy by probing the presence of adversarial patches, which allows for more comprehensive detection. More importantly, our method introduces randomness, which significantly outperforms UDF when confronted with adaptive attacks (see \S\ref{subsec:eval:adaptive}).
}
\revend{D}{4}

% \textbf{Enhancing Object Detectors.}
% In addition to removing adversarial patches, enhancing object detection models \cite{DongRobustDet}\cite{DetectorGuard}\cite{hein2017formal}\cite{shafahi2019adversarial} is also used to defend against adversarial patch attacks. Dong et al. \cite{DongRobustDet} retrains various types of object detectors with adversarial examples to enable the object detection models to learn common features between adversarial and benign samples. However, this reduces the detector's efficiency, and choosing an appropriate adversarial example is a non-trivial problem. DetectorGuard \cite{DetectorGuard} introduces an additional robust image classifier named Objectness Predictor to explain the high objectness objects detected by the base detector. If any object can not be explained means the adversarial patch attack is detected.  However,  DetectorGuard is model-specific and may lead to false positives.
% Enhancing object detection requires retraining or finetuning the object detection model or introducing an auxiliary model. Retraining or finetuning the object detection model will reduce the detection performance, and picking a suitable training set is hard.

%% file: Conclusion.tex
\section{Conclusion}

This paper proposes a novel defense method for detecting adversarial patch attacks with defensive patches.
Two types of defensive patches, named \textit{canary} and \textit{woodpecker}, are imported into the input image to proactively probe or counteract potential adversarial patches. 
The proposed method provides an effective and efficient solution for adversarial patch attacks in a completely new way.
% A comprehensive experiment demonstrated that our method can achieve very high precision and recall, even facing unknown attack methods. 
% \newreplace{
A comprehensive experiment demonstrated that our method can achieve high performance and outperform existing defense methods, even facing unknown attack methods.
% }
The time overhead is also limited. 
Furthermore, we design randomized canary and woodpecker injection patterns to defend against defense-aware attacks. 
The experiment demonstrated that the proposed method is capable of withstanding adaptive attacks. 
We believe that our method can be applied to various object detectors, and is practical in 
real-world scenarios.

%% file: Appendix.tex
% \newpage

\begin{figure*}[!t]
  \centering
  \includegraphics[width=\textwidth]{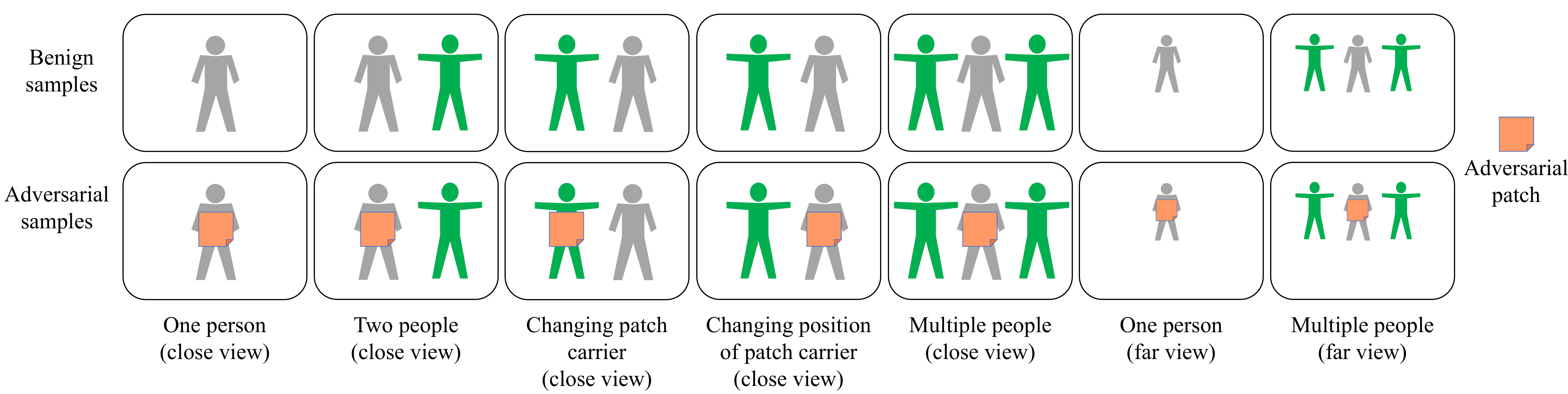}
\caption{Various situations for collecting physical-world benign and adversarial samples.}
\label{fig-get-dataset}
\end{figure*}

% \begin{figure*}[!t]
%     \centering
%     \includegraphics[width=\textwidth]{figs/effective_analysis_cawd.pdf}
%     \caption{Evaluating canary and woodpecker in physical-world scenarios. Red boxed images contain defensive patches and are detected. (a)-(f) present examples in different indoor scenarios. (g)-(l) present examples in different outdoor scenarios.}
%     \label{fig-effective-analysis-cawd}
% \end{figure*}  

\appendix

\renewcommand\thesubsectiondis{\Alph{subsection}.}

\subsection{ Physical-world Dataset}
\label{appendx:physicalworlddataset}

We applied the four attack methods to generate adversarial patches and printed them out. 
As shown in Figure~\ref{fig-get-dataset}, 
we then took photos to get physical-world adversarial images (with a printed patch held by a person) and benign samples (without patches) in 14 situations. % (half indoors and half outdoors).
Each situation was further composed of the same number of indoor and outdoor scenarios. In total, we got 1,960 adversarial samples and 1,960 benign samples for the evaluation of effectiveness (\S\ref{EffectivenessAnalysis}).

\input{tables/table-dfpatch-class-position}

\begin{figure*}[!t]
    \centering
    \includegraphics[width=\textwidth]{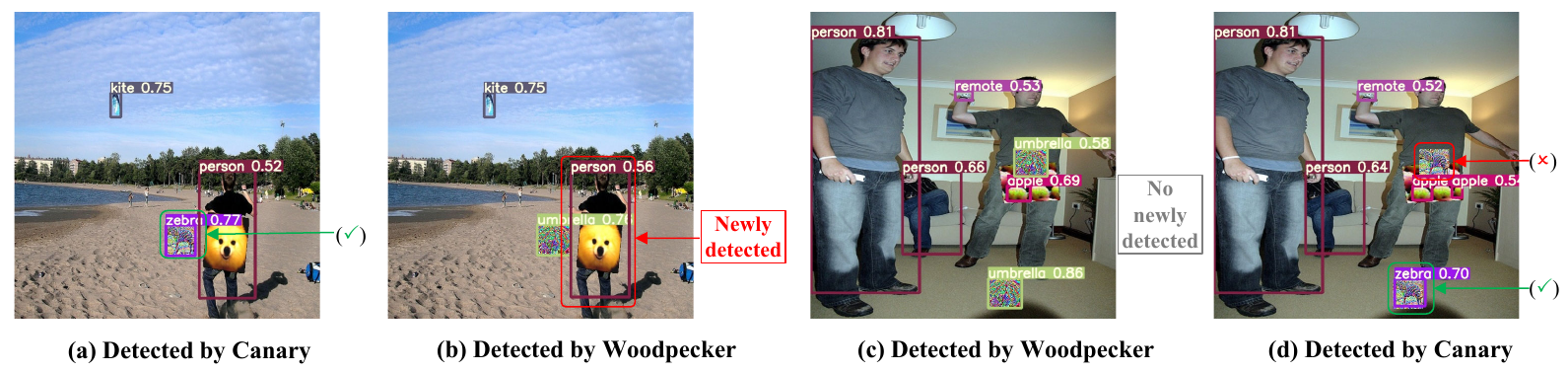}
    \caption{False negatives using (a) canary and (c) woodpecker in adversarial examples. \mdseries Note that, applying the other kind of defensive patch can effectively detect the attacks, i.e., (b) and (d).}
    \label{fig-false-negative-cawd}
\end{figure*}

\begin{figure}[!t]
\centering
\includegraphics[width=\linewidth]{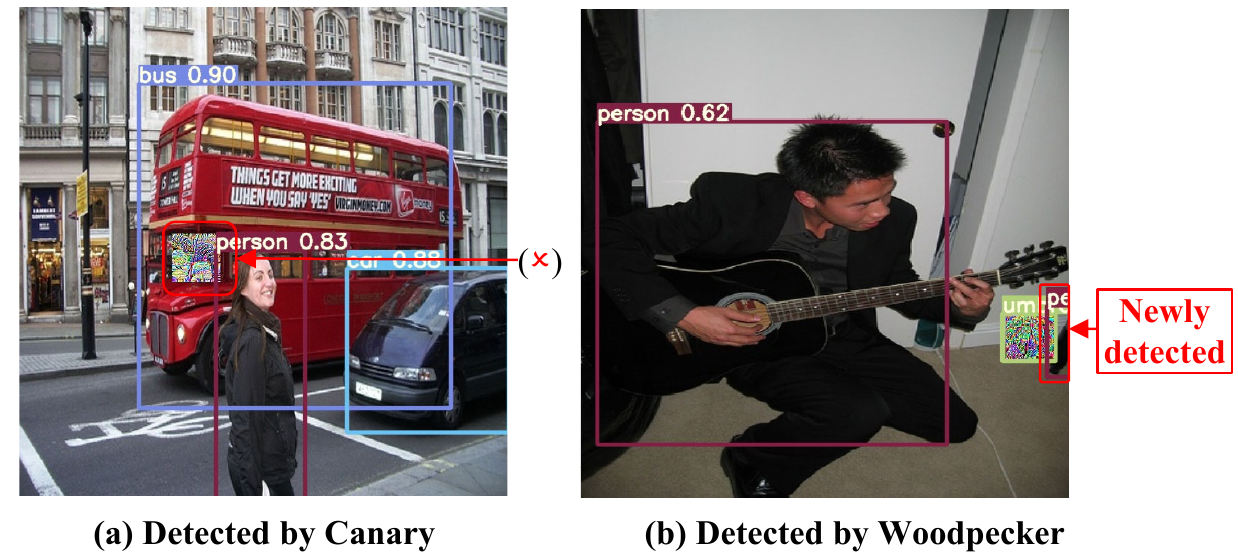}
\caption{False positives using canary and woodpecker in benign samples.}
\label{fig-false-positive-cawd}
\end{figure}

\subsection{False Negative and False Positive Cases.} 
\label{appendix:FN FP}
% Our experiments have shown that Mode \#1 (canary only) or Mode \#2 (woodpecker only) may result in FNs and FPs. 
% We present illustrative examples below.

Figure~\ref{fig-false-negative-cawd}(a) shows an example where the introduced canary is detected in an adversarial image, i.e., the adversarial patch attack is not detected, as the perturbation effect of the adversarial patch may be not enough to affect the canary, resulting an FN. 
Similarly, in Figure~\ref{fig-false-negative-cawd}(c), we fail to report the attack with only the woodpecker patch, when the woodpecker may occasionally fail to influence the adversarial patch. 
% In specific scenarios, the perturbation effect of the adversarial patch may be not enough to affect the canary, so the canary can be correctly identified (Figure~\ref{fig-false-negative-cawd} (a)), thereby resulting in a false negative.
% Additionally, the woodpecker may occasionally fail to influence the adversarial patch, leading to a failure to successfully restore the attacked target, as illustrated in Figure~\ref{fig-false-negative-cawd} (c).
Fortunately, complementing the defense with the other technique will solve the problem. 
Applying woodpecker as in Figure~\ref{fig-false-negative-cawd}(b) helps alert the attack, and the canary in Figure~\ref{fig-false-negative-cawd}(d) is disrupted by the adversarial patch.

Figure~\ref{fig-false-positive-cawd} shows two examples that canary and woodpecker produce FPs. 
% As a fragile object, canary may not be accurately detected  in scenarios with complex backgrounds due to background interference.
In Figure~\ref{fig-false-positive-cawd}(a), the introduced canary is not identified accurately due to the complex background in a benign sample, leading to a false alarm about a non-existent attack. 
Woodpecker may occasionally misidentify the content near the image borders as attacked targets.
In Figure~\ref{fig-false-positive-cawd}(b), a small portion near the border of the image was incorrectly identified as the attacked target after introducing woodpecker.
% Fortunately, the above scenarios are exceedingly rare, so 
% employing defensive patches remains a viable strategy for detecting adversarial patch attacks.
We believe that using a more diverse training set to generate canaries and woodpeckers can effectively reduce false positives.

\subsection{Effectiveness of Different Initial Classes and Placement Positions}
\label{appendx:initialplacement}

We investigated how the object classes and positions of the initial defensive patches may affect the performance of defensive patches, targeting AdvPatch on YOLOv8 in the three public datasets.
Three different initial categories (\textit{elephant}, \textit{zebra} and \textit{giraffe}) were selected and trained to generate five canaries for each class, corresponding to five positions. 
In addition, for each of the five positions, we generate a woodpecker. 
% 
% \textcolor{red}{\hl{\bf [To update according to Table 4.]}}
The results are shown in Table~\ref{tab-dfpatch-class-position}.  
It can be seen that the initial classes or placement positions do not have great impact on the detection performance. 

%% file: tables/table-dfpatch-class-position.tex
\begin{table}
\centering
\caption{
\ifshepherd
\revbegin{B}{7}
\sethlcolor{yellow}\hl{The performance of Mode \#1 and Mode \#2 in different classes and positions.} 
\revend{B}{7}
\else {The performance of Mode \#1 and Mode \#2 in different classes and positions.}
\fi 
}

\begin{tabular}{ccccccc} 
\hline
\begin{tabular}[c]{@{}c@{}}Defense\\ Patch\end{tabular}  & Dataset & Center & Up & Down  & Left  & Right  \\ 
\hline
\multirow{3}{*}{\begin{tabular}[c]{@{}c@{}}Canary\\ (\textit{elephant})\end{tabular}} & VOC07 & 0.936  & 0.947 & 0.950 & 0.963 & 0.935  \\
 & COCO & 0.949  & 0.969 & 0.957 & 0.967 & 0.936  \\
 & Inria & 0.988  & 0.984 & 0.988 & 0.980 & 0.937  \\ 
\hline
\multirow{3}{*}{\begin{tabular}[c]{@{}c@{}}Canary\\ (\textit{zebra})\end{tabular}} & VOC07 & 0.971  & 0.967 & 0.967 & 0.965 & 0.966  \\
 & COCO & 0.975  & 0.965 & 0.960 & 0.969 & 0.963  \\
 & Inria & 0.980  & 0.988 & 0.992 & 0.988 & 0.984  \\ 
\hline
\multirow{3}{*}{\begin{tabular}[c]{@{}c@{}}Canary\\ (\textit{giraffe})\end{tabular}}  & VOC07 & 0.971  & 0.946 & 0.957 & 0.969 & 0.975  \\
 & COCO & 0.971  & 0.955 & 0.959 & 0.973 & 0.963  \\
 & Inria & 0.992  & 0.984 & 0.980 & 0.980 & 0.992  \\ 
\hline
\multirow{3}{*}{Woodpecker} & VOC07 & 0.880  & 0.877 & 0.901 & 0.856 & 0.918  \\
 & COCO & 0.877  & 0.889 & 0.868 & 0.850 & 0.919  \\
 & Inria & 0.936  & 0.858 & 0.953 & 0.902 & 0.971  \\
\hline
\end{tabular}

\label{tab-dfpatch-class-position}
\end{table}

%% file: Meta-review.tex
\newpage % The Meta-Review should at least start on a new column

% Use \appendices and not \appendix due to IEEEtran.cls quirks
% \appendices % if not used earlier

\subsection{Meta-Review}

The following meta-review was prepared by the program committee for the 2025
IEEE Symposium on Security and Privacy (S\&P) as part of the review process as
detailed in the call for papers.

\subsubsection{Summary}
This paper introduces a new defense method against adversarial patch attacks. This method injects defensive patches, consisting of canary and woodpecker objects, that are meant to probe and counteract potential adversarial patches. By detecting the presence of these patches near the bounding boxes of potential adversarial patches, the defender can detect if an attack has occurred in the region.

\subsubsection{Scientific Contributions}
\begin{itemize}
\item Addresses a Long-Known Issue.
\item Creates a New Tool to Enable Future Science.
\item Provides a Valuable Step Forward in an Established Field.
\end{itemize}

\subsubsection{Reasons for Acceptance}
\begin{enumerate}
\item A long-known issue is addressed and a valuable step forward is made in an established field. Machine learning models have been repeatedly demonstrated to be vulnerable to a plethora of trustworthiness and security issues. This paper aims to defend vulnerable models and achieves competitive results even against adaptive attacks.
\item This paper creates a new tool to enable future science. The authors introduce a defense method that does not require additional retraining or modification of model weights. Given the significant burden on time that training machine learning models requires, a defense method that works with out of the box models is more likely to be implemented.
\end{enumerate}

% \subsection{Noteworthy Concerns} % Exclude if your meta-review does not have noteworthy concerns
% \begin{enumerate} % Enumerate environment is not necessary if there is only one
% \item text
% \item text
% \end{enumerate}

% \section{Response to the Meta-Review} % Optional